\def\arxivpreprint{}
\documentclass{article} 
\usepackage{iclr2027_conference,times}

\usepackage{amsmath,amsfonts,bm}

\def\eqref#1{equation~\ref{#1}}

\def\1{\bm{1}}

\DeclareMathAlphabet{\mathsfit}{\encodingdefault}{\sfdefault}{m}{sl}
\SetMathAlphabet{\mathsfit}{bold}{\encodingdefault}{\sfdefault}{bx}{n}

\usepackage{amssymb}
\usepackage{booktabs}
\usepackage{graphicx}
\usepackage{multirow}
\usepackage[caption=false,labelformat=simple]{subfig}
\renewcommand{\thesubfigure}{(\alph{subfigure})}
\usepackage{hyperref}
\usepackage{url}
\usepackage{xspace}

\makeatletter
\AddToHook{env/tabular/begin}{\let\input\@@input}
\makeatother

\newcommand{\method}{Align-VAE\xspace}
\newcommand{\groupedtablehead}[2]{%
  \multirow[c]{#1}{*}%
    [-\dimexpr(\aboverulesep+\cmidrulewidth+\belowrulesep)/2\relax]{#2}}
\DeclareRobustCommand{\methodpanelnumber}[1]{\textcircled{\scriptsize #1}}
\DeclareRobustCommand{\ablationvariant}[1]{%
  \begingroup\hypersetup{hidelinks}%
  \hyperlink{ablation-definitions}{#1}\endgroup}

\title{\method: Aligning Attention and Geometry in VecSet VAEs for 3D Reconstruction}

\ifdefined\arxivpreprint
  \newcommand{\arxivauthors}{Huaizhi Qu, Mingxin Yang, Xintong Han, Fangneng Zhan, Chunchao Guo}
\author{%
  Huaizhi Qu\textsuperscript{1,2}\quad
  Mingxin Yang\textsuperscript{2}\quad
  Xintong Han\textsuperscript{2}\quad
  Fangneng Zhan\textsuperscript{1,\textdagger}\quad
  Chunchao Guo\textsuperscript{2,\textdagger}\\
  \textnormal{\textsuperscript{1}The Hong Kong University of Science and Technology}\quad
  \textnormal{\textsuperscript{2}Tencent Hunyuan}\\
  \textnormal{\textsuperscript{\textdagger}Corresponding authors.}%
}

  \iclrfinalcopy
  \hypersetup{
    hidelinks,
    pdftitle={Align-VAE: Aligning Attention and Geometry in VecSet VAEs for 3D Reconstruction},
    pdfauthor={\arxivauthors}
  }
\else
  \author{Anonymous Authors}
\fi

\begin{document}

\maketitle
\ifdefined\arxivpreprint
  \lhead{Preprint}
\fi

\begin{abstract}
\vspace{-4.8pt}
Set-based variational autoencoders (VecSet VAEs) represent 3D shapes as
compact sets of latent tokens. Although each token originates from a query
at a sampled surface location, whether these latents truly align with
geometry remains unclear. We investigate this alignment and identify four
issues: (i)~attention-sink-like tokens that attract queries from across the
shape; (ii)~spatially periodic repetitions in decoder attention responses;
(iii)~mismatched normal information in encoder and decoder query features;
and (iv)~missing supervision of the intrinsic relation between signed distance
fields and surface normals. To address these issues, we introduce \method
with four corresponding modifications: (i)~a shared bank of learnable decoder
tokens provides a workspace for global computation; (ii)~distance-biased
cross-attention guides retrieval toward nearby predicted spatial anchors;
(iii)~normal-free encoder queries align the two stages' access to normal
information while retaining normals in the encoder keys and values; and
(iv)~a screened Poisson-inspired regularizer supervises surface values and
normal-direction derivatives. These changes align latent
retrieval and field prediction with geometry without enlarging the
per-shape code.
On the ABO dataset, \method reduces Chamfer distance by 26.6\%
compared to the standard VecSet VAE.
\end{abstract}

\begin{figure}[!ht]
\centering
\vspace{-4pt}
\includegraphics[width=.94\textwidth]{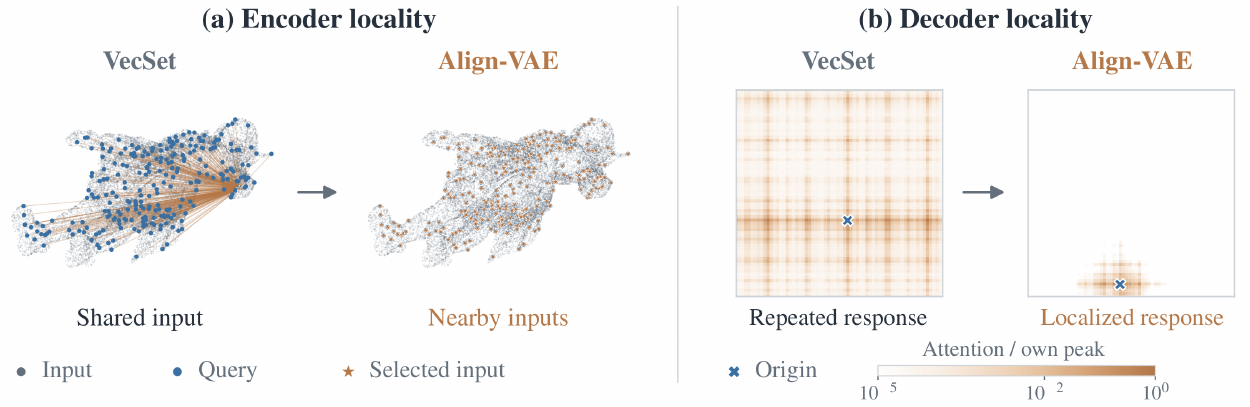}
\caption{\textbf{Improving local responses in VecSet VAEs.}
(a) Nearby encoder selections replace a shared input.
(b) Decoder responses become localized.
The models share an input shape and 8192 stored tokens.
Lines show most-attended inputs; crosses mark FPS origins.
Each slice shows the median token by surface attention, through its
volume peak and normalized by that peak.}
\label{fig:teaser}
\vspace{-4pt}
\end{figure}

\vspace{-4.8pt}
\section{Introduction}
\label{sec:introduction}

\vspace{-4.8pt}
Compact 3D representations provide an interface between detailed geometry
and latent generative models. VecSet variational autoencoders (VecSet VAEs) compress a
surface into a fixed number of latent tokens and reconstruct it through
an implicit field queried at arbitrary spatial locations
\citep{zhang20233dshape2vecset}. Their token budget can be chosen
independently of the resolution at which the field is decoded, making
them attractive for scalable shape modeling
\citep{zhang2024clay,li2025triposg,hunyuan3d2025hunyuan3d}.
Increasing output sampling density cannot recover detail
that the code fails to preserve or the decoder fails to retrieve.
Figure~\ref{fig:teaser} previews how \method improves local responses
during encoding and decoding while retaining the stored code budget.

\vspace{-4.8pt}
In a point-query VecSet encoder, each latent originates from a query at a
sampled surface point. This assigns a spatial origin, but does it also
establish a spatial role? Encoder and decoder
cross-attention retrieve features by content
similarity, while self-attention mixes information across tokens. None
of these operations requires a token to describe its originating
neighborhood. Global context helps describe a coherent shape, while local
field queries need spatially precise evidence. We hypothesize that sharing
limited latent capacity across these roles can leave spatially seeded
latents poorly aligned with the local geometry they are meant to represent.

\vspace{-4.8pt}
Our profiling of a standard VecSet VAE reveals gaps between spatial
origins and geometric roles. Encoder queries often favor shared inputs,
while decoder attention concentrates on a small subset of latents
(Figures~\ref{fig:encoder-locality-statistics} and~\ref{fig:decoder-mass}).
Individual latents can respond at separated locations, and decoded field
gradients deviate from surface normals (Figures~\ref{fig:response-slices}
and~\ref{fig:baseline-field-normal}). Encoder queries also use observed
normals that arbitrary decoder queries lack.
Section~\ref{sec:alignment-analysis} explains how these diagnostics
motivate changes to local encoding, spatial retrieval, and field supervision.

\vspace{-4.8pt}
We introduce \method to address these issues while retaining the stored
VecSet code. A shared bank of learnable decoder tokens supplies a
workspace for global computation. Predicted spatial anchors and a
distance bias guide retrieval from the remaining shape tokens toward
nearby geometry. Normal-free encoder queries remove the normal-information
mismatch while keeping observed normals in the support keys and values.
Finally, a regularizer inspired by screened Poisson
reconstruction~\citep{kazhdan2013screened} constrains surface values
and variation along the observed normals. Together, these changes govern
how local evidence is encoded, retrieved, and translated into a geometric
field with spatial and orientation cues.

\vspace{-4.8pt}
Our contributions are threefold:
\vspace{-4.8pt}
\begin{itemize}
    \vspace{-1.6pt}
    \item \textbf{Detailed analysis of VecSet VAEs.} We examine attention
    allocation, spatial retrieval, query information, and field orientation
    to characterize the geometric roles of spatially seeded latents
    throughout encoding and decoding.
    \vspace{-1.1pt}
    \item \textbf{Targeted improvements.} We introduce \method, combining
    a shared decoder bank, distance-biased retrieval, normal-free queries,
    and normal supervision without enlarging the stored per-shape code.
    \vspace{-1.1pt}
    \item \textbf{Extensive evaluation.} We evaluate \method through
    reconstruction benchmarks, component ablations, and geometric diagnostics.
    On the ABO dataset, we achieve a 26.6\% reduction in Chamfer distance
    compared to the standard VecSet VAE.
\end{itemize}

\vspace{-10.1pt}
\section{Related Work}
\label{sec:related_work}

\vspace{-5.7pt}
\paragraph{Compact 3D shape latents.}
3DILG~\citep{zhang20223dilg} associates latent features with irregular
spatial samples, while 3DShape2VecSet~\citep{zhang20233dshape2vecset}
uses attention to connect a compact latent set to continuous field
queries. Michelangelo~\citep{zhao2023michelangelo} aligns shape latents
with image and text features, and CLAY~\citep{zhang2024clay},
TripoSG~\citep{li2025triposg}, and
HY3D-2.1~\citep{hunyuan3d2025hunyuan3d} scale set-based shape modeling.
Dora~\citep{chen2025dora} studies sampling and evaluation for 3D VAEs.
LATTICE~\citep{lai2025lattice} anchors latents to a coarse voxel grid;
MSVS-VAE~\citep{hao2026msvsvae} combines anchored latent hierarchies with
local decoding.
We analyze the geometric role of VecSet attention.

\vspace{-5.9pt}
\paragraph{Spatially structured representations.}
XCube~\citep{ren2024xcube} and TRELLIS~\citep{xiang2025structured}
organize 3D features on sparse spatial grids. Direct3D-S2 and
Sparc3D model detailed shapes with sparse
representations~\citep{wu2025direct3d,li2025sparc3d}.
These methods provide explicit spatial organization; VecSet methods
decouple the stored token count from a dense decoding grid.
We report accuracy, storage, and decoding cost separately because token
widths and layouts differ.

\vspace{-5.9pt}
\paragraph{Computational workspace and oriented geometry.}
Registers provide additional tokens for internal computation in vision
transformers~\citep{darcet2024registers}. We test this workspace
interpretation in a 3D VAE by separating an unanchored decoder bank from
spatially addressed shape tokens.
Poisson and screened Poisson reconstruction connect oriented surface
samples to an implicit field~\citep{kazhdan2006poisson,kazhdan2013screened}.
Our regularizer draws on surface interpolation and normal-direction
constraints to supervise the learned field during training. Inference
evaluates this field without solving a Poisson reconstruction system.

\ifdefined\arxivpreprint
  \newpage
\fi
\vspace{-5.9pt}
\section{Preliminaries}
\label{sec:preliminaries}
\label{sec:problem-setup}
\label{sec:vecset-backbone}

\vspace{-5.7pt}
An oriented point set $\mathcal P=\{(\mathbf x_j,\mathbf n_j,s_j)\}_{j=1}^N$
contains normalized surface positions, outward unit normals, and scalar
sharp-edge labels. Farthest-point sampling (FPS) selects $M$ centers
$\mathcal C=\{\mathbf c_i\}_{i=1}^M$. Each center supplies one query
over the $N$ support points. Encoder cross-attention and self-attention
yield one latent token of width $C_z$ per center. The autoencoder maps are
\begin{equation}
    \bm Z=\operatorname{Enc}(\mathcal P,\mathcal C)\in\mathbb R^{M\times C_z},
    \qquad
    \hat f(\mathbf q;\bm Z)=\operatorname{Dec}(\mathbf q,\bm Z).
    \label{eq:prelim-autoencoder}
\end{equation}
The decoder refines $\bm Z$ with self-attention, then reads the tokens
through cross-attention from an embedded spatial query $\mathbf q$.
The zero level set of $\hat f$ defines the reconstructed surface.
The target is a normalized truncated signed distance field (TSDF):
\begin{equation}
    y(\mathbf q)=r\,\operatorname{clip}
       (f_{\mathrm{sdf}}(\mathbf q),-1/r,1/r),
    \label{eq:tsdf-target}
\end{equation}
where $r>0$ is the inverse truncation half-width and the SDF is positive
outside. At a smooth surface point, $y(\mathbf x)=0$ and
$\mathbf n^\top\nabla y(\mathbf x)=r$. The token count $M$ fixes the
stored code budget independently of the number of decoder queries.

\vspace{-4.8pt}
\section{Do Spatial Queries Produce Geometrically Aligned Latents?}
\label{sec:alignment-analysis}

\begin{figure}[tbp]
\centering
\begin{minipage}[t]{.43\textwidth}
\centering
\subfloat[Encoder input selection\label{fig:encoder-locality}]{%
  \includegraphics[page=1,width=\linewidth]{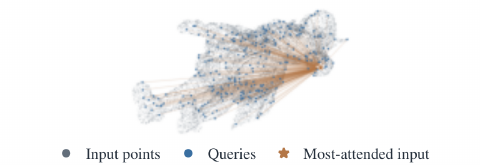}}
\end{minipage}\hspace{.01\textwidth}%
\begin{minipage}[t]{.43\textwidth}
\centering
\subfloat[Input selection rates\label{fig:encoder-locality-statistics}]{%
  \includegraphics[page=2,width=\linewidth]{figures/rasterized/attention_diagnostics}}
\end{minipage}

\begin{minipage}[t]{.43\textwidth}
\centering
\subfloat[Cumulative decoder attention\label{fig:decoder-mass}]{%
  \includegraphics[page=3,width=\linewidth]{figures/rasterized/attention_diagnostics}}
\end{minipage}\hspace{.01\textwidth}%
\begin{minipage}[t]{.43\textwidth}
\centering
\subfloat[Surface response maps\label{fig:decoder-spatial-readout}]{%
  \includegraphics[page=4,width=\linewidth]{figures/rasterized/attention_diagnostics}}
\end{minipage}
\vspace{-4pt}
\caption{Attention allocation in a standard VecSet VAE.
Local selection uses 64 nearest inputs; the dotted line denotes equal
allocation. Surface maps (attention ranks 1, 10, median) use each token's
surface peak for log-scale normalization; crosses mark FPS origins.
Bars and curves show shape means and 95\% intervals; (a, d) share one example.}
\label{fig:attention-diagnostics}
\vspace{-15pt}
\end{figure}

\vspace{-4.8pt}
Figure~\ref{fig:attention-diagnostics} profiles attention in a standard
VecSet VAE. Each encoder query
starts at an FPS surface point and gathers information from the input
cloud into a latent token. The decoder then reads these tokens from
arbitrary 3D query locations to predict an implicit field. Our profiles
examine where information is collected, where it is retrieved, and how
it determines surface orientation. Appendix~\ref{sec:profiling-guide}
illustrates the profiles that motivate the four components of \method.

\vspace{-4.8pt}
\subsection{Nonlocal Encoder Selection and Uneven Decoder Attention}
\label{sec:sink-observations}

\vspace{-4.8pt}
\paragraph{Encoder input selection.}
To test whether distributed query origins yield local reads, we link
each of the 8192 FPS encoder queries to the input point with its highest
head-averaged attention weight (Figure~\ref{fig:encoder-locality}).
Figure~\ref{fig:encoder-locality-statistics} reports the fraction selecting
one of its 64 nearest inputs in Euclidean distance and the fraction
selecting the shape's most-selected input. The mean rates are 0.07\%
and 94.1\%, respectively: dominant selections largely converge on a
shared point despite distributed FPS origins. This behavior resembles
attention sinks in LLMs~\citep{xiao2024efficient} and the use of spatial
tokens for global computation in ViTs~\citep{darcet2024registers}.
A per-head analysis also finds two heads with mean local-selection rates
above 99\% (Appendix~\ref{sec:measured-figure1}), showing that shared
selection coexists with local information collection.

\vspace{-4.8pt}
\paragraph{Decoder attention concentration.}
We next measure how field prediction uses the latent budget. A token's
\emph{attention mass} is its average attention weight over decoder
queries, representing its share of the decoder's attention. We evaluate
queries throughout the volume and on the input surface separately.
Figure~\ref{fig:decoder-mass} ranks tokens by this mass and plots their
cumulative share. The top 1\% (82 of 8192 tokens) receive 19.5\% of
volume-query attention and 9.8\% of surface-query attention, compared
with approximately 1\% under equal allocation. This concentration
extends the sink-like pattern to latent retrieval and suggests competition
between shared computation and local geometric storage. This motivates
a shared decoder bank: learnable tokens provide a workspace for global
computation, so shape tokens can focus on local geometry without enlarging
the stored code.

\subsection{Spatial Repetition in Decoder Cross-Attention}
\label{sec:periodic-observation}

\vspace{-4.8pt}
\paragraph{Surface response maps.}
After measuring how much attention each latent receives, we examine
\emph{where} it is read. A token's spatial response is the attention
assigned to it as the decoder query moves through 3D space.
Figure~\ref{fig:decoder-spatial-readout} visualizes this response on the
surface for tokens ranked first, tenth, and median by surface attention
mass. Each map scales colors by that token's own peak response, while
crosses mark the originating FPS points. Comparing the colored regions
with these origins reveals the spatial extent and displacement of retrieval.
These maps show diffuse, smoothly varying responses over broad surface
regions, rather than compact peaks around their FPS origins.

\begin{figure}[tbp]
\centering
\begin{minipage}[t]{.43\textwidth}
\centering
\subfloat[Maximum-response positions\label{fig:response-maxima}]{%
  \includegraphics[page=1,width=\linewidth]{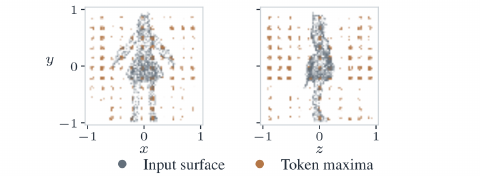}}
\end{minipage}\hspace{.01\textwidth}%
\begin{minipage}[t]{.43\textwidth}
\centering
\subfloat[Token-response slices\label{fig:response-slices}]{%
  \includegraphics[page=2,width=\linewidth]{figures/rasterized/spatial_profiles}}
\end{minipage}

\begin{minipage}[t]{.43\textwidth}
\centering
\subfloat[Response trace and shift similarity\label{fig:response-traces}]{%
  \includegraphics[page=3,width=\linewidth]{figures/rasterized/spatial_profiles}}
\end{minipage}\hspace{.01\textwidth}%
\begin{minipage}[t]{.43\textwidth}
\centering
\subfloat[Field-normal angular error\label{fig:baseline-field-normal}]{%
  \includegraphics[page=4,width=\linewidth]{figures/rasterized/spatial_profiles}}
\end{minipage}
\vspace{-4pt}
\caption{Spatial responses and field orientation in standard VecSet.
Shape and token ranks match Figure~\ref{fig:decoder-spatial-readout}.
Slices are normalized by each token's full-volume peak; crosses mark projected FPS origins.
The dashed line gives the trace in (c), alongside shift similarity
$C_{\mathrm{rep}}$. Normal maps saturate at $90^\circ$.
Summaries show shape means and 95\% intervals.}
\label{fig:periodic-responses}
\vspace{-15pt}
\end{figure}

\vspace{-4.8pt}
\paragraph{Response slices and traces.}
Figure~\ref{fig:periodic-responses} extends this view into the volume.
Figure~\ref{fig:response-maxima} marks the location where each token
receives its strongest attention. Figure~\ref{fig:response-slices}
shows the same selected tokens' responses on 2D slices, and
Figure~\ref{fig:response-traces} follows a straight line through a
response maximum. The separated peaks show that the same token is
retrieved at multiple locations, including regions far from its FPS
origin. This is a retrieval-locality mismatch: a token seeded at one
surface neighborhood is read as evidence for distant field queries.

\vspace{-4.8pt}
\paragraph{Similarity across spatial shifts.}
To summarize the response structure, we shift each one-dimensional
attention curve along its spatial axis and correlate the shifted curve
with the original. We denote the highest correlation over a fixed set
of nonzero shifts by $C_{\mathrm{rep}}$: larger values indicate more
similar responses at separated locations. Mean scores are 0.475, 0.518,
and 0.274 for the first, tenth, and median-ranked tokens, respectively
(Figure~\ref{fig:response-traces}). Together with the separated response
peaks, this profile motivates an explicit spatial reference for decoder
retrieval. We predict an anchor for each shape token and bias attention
toward anchors near the query, so retrieval uses both feature similarity
and geometric proximity.

\vspace{-4.8pt}
\subsection{Query-Normal Asymmetry}
\label{sec:remaining-gaps}

\vspace{-4.8pt}
The encoder--decoder interface introduces a second retrieval mismatch.
Encoder queries originate at observed surface points and include their
normals. Decoder queries can lie anywhere in space, where a surface
normal is unavailable. The encoder can therefore organize information
using an orientation cue that the decoder cannot supply when retrieving
it. This motivates normal-free encoder queries: removing normals from
the query features makes both stages locate evidence without an input
normal, while retaining normals in encoder keys and values preserves
the observed orientation information. Perturbing only query normals,
with input positions and support features fixed, provides a direct test
of this dependence through attention and reconstruction
(Table~\ref{tab:query_normal}).

\vspace{-4.8pt}
\subsection{Field--Normal Misalignment}
\label{sec:field-normal-gap}

\vspace{-4.8pt}
Geometric alignment also depends on how retrieved features become a
field. At the zero level set, the field gradient determines the surface
normal, so correct orientation requires the predicted field to vary in
the appropriate spatial direction. At observed surface points, we query
small offsets along each coordinate axis, estimate the field gradient,
and measure its angular difference from the observed outward normal.
Figure~\ref{fig:baseline-field-normal} shows a mean error of $13.52^\circ$.
This discrepancy exposes an orientation gap in supervision based on field
values alone. It motivates coupling field regression with surface and
normal-direction constraints: the field should vanish on the observed
surface and change along its normals according to the signed-distance
target. Our screened Poisson-inspired regularizer imposes these two
geometric conditions alongside the attention improvements.

\vspace{-4.8pt}
\section{Method}
\label{sec:method}

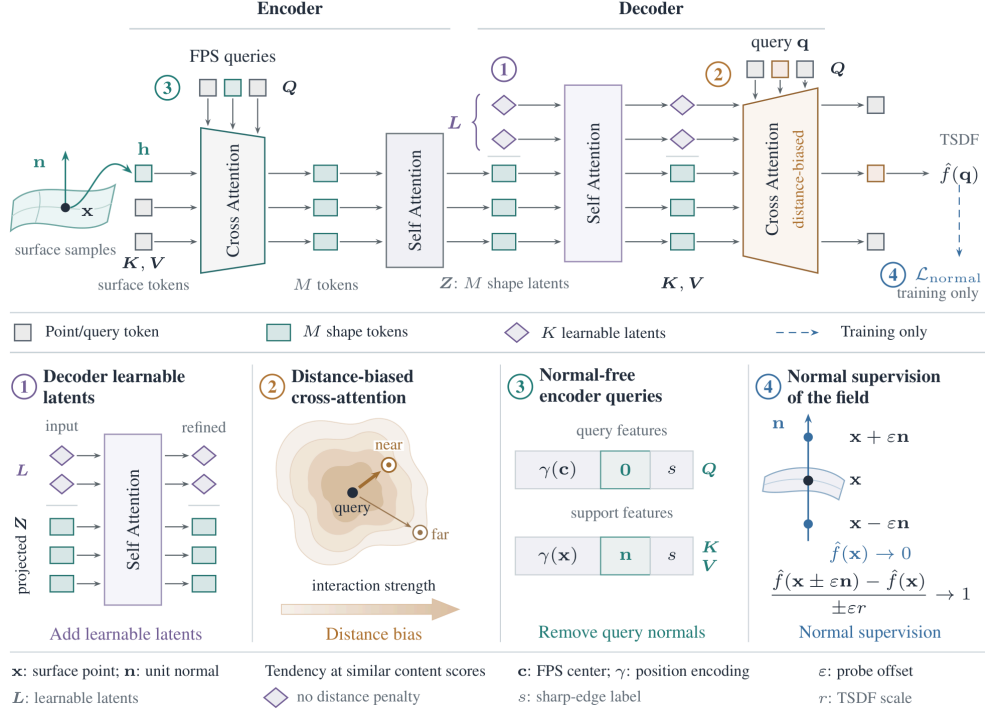
\begin{figure}[t]
\centering
\renewcommand{\thesubfigure}{\protect\methodpanelnumber{\arabic{subfigure}}}
\setlength{\unitlength}{\textwidth}
\begin{picture}(0,0)
  \put(.02,.33){\subfloat{\label{fig:method-bank}\makebox[0pt]{}}}
  \put(.27,.33){\subfloat{\label{fig:method-distance-bias}\makebox[0pt]{}}}
  \put(.52,.33){\subfloat{\label{fig:method-normal-free-queries}\makebox[0pt]{}}}
  \put(.77,.33){\subfloat{\label{fig:method-normal-supervision}\makebox[0pt]{}}}
\end{picture}%
\includegraphics[width=.94\textwidth]{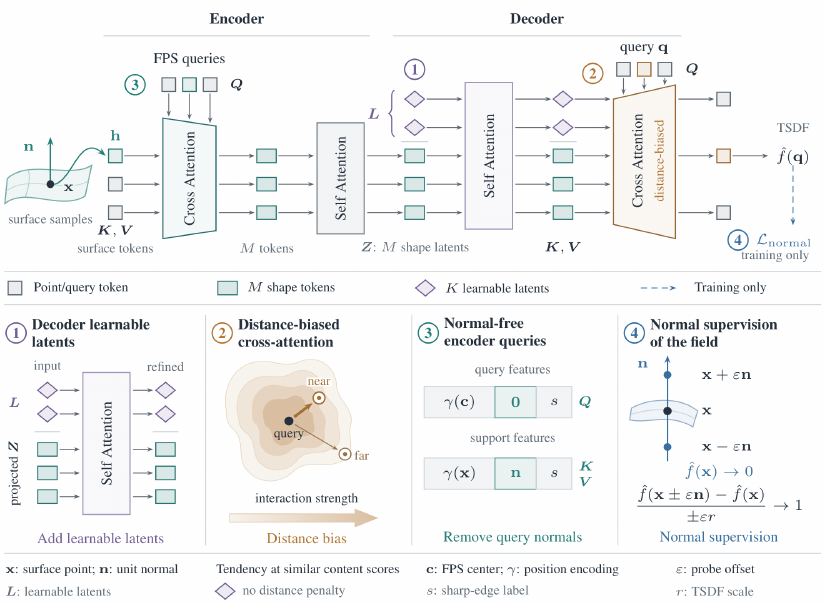}
\vspace{-4pt}
\caption{\method overview. Encoder cross-attention and self-attention
produce the $M$-token code. The decoder prepends $K$ shared learnable
tokens before self-attention and retrieves shape tokens with a distance
bias. Circled regions show the decoder bank, distance bias, normal-free
queries, and normal supervision, respectively. The field receives surface
and normal-direction constraints; color bands depict the spatial prior.}
\label{fig:method}
\vspace{-15pt}
\end{figure}

\vspace{-4.8pt}
\method retains the $M$-token variational code and changes how geometric
information is collected, retrieved, and supervised. Figure~\ref{fig:method}
summarizes the architecture. Each component addresses a baseline
profile from Section~\ref{sec:alignment-analysis};
Figure~\ref{fig:component-profiles} compares the complete models at 8192 latents.
Implementation details and training settings appear in
Appendix~\ref{sec:method-details}.

\vspace{-4.8pt}
\subsection{A Decoder Workspace Alongside Shape Tokens}
\label{sec:learnable-latents}

\vspace{-4.8pt}
The decoder mass and spatial profiles in
Figures~\ref{fig:decoder-mass} and~\ref{fig:decoder-spatial-readout}
motivate a register-like workspace for global computation~\citep{darcet2024registers}
(Figure~\ref{fig:method-bank}). A learned bank
$\bm L\in\mathbb R^{K\times d}$, shared across shapes, is concatenated
with the projected code:
\begin{equation}
    \widetilde{\bm Z}=T_{\mathrm{dec}}
       ([\bm L\mid P_{\mathrm{dec}}(\bm Z)]).
    \label{eq:main-bank}
\end{equation}
Here $P_{\mathrm{dec}}$ projects tokens to width $d$,
$[\cdot\mid\cdot]$ concatenates them, and $T_{\mathrm{dec}}$ is the
decoder self-attention stack. This interaction makes the bank
shape-dependent. Spatial queries read both populations: shape tokens
can support local retrieval, while the bank supplies context without
anchors or distance penalties. Attention-mass analysis tests this
allocation of capacity. The bank adds $Kd$ shared parameters and $K$
decoder tokens, leaving the stored code and KL term unchanged.

\vspace{-4.8pt}
\paragraph{Profiling the bank.}
Figure~\ref{fig:component-profiles}(a) compares the concentration of surface-query
attention on shape tokens in VecSet and \method.
Table~\ref{tab:attention_profile} separates bank share $G_{\mathrm{bank}}$
from shape-token read distances and spreads in the cumulative variants.

\subsection{Normal-Free Encoder Queries}
\label{sec:position-only-queries}

\vspace{-4.8pt}
To remove the query-normal mismatch, we mask normal channels only in
the encoder query branch (Figure~\ref{fig:method-normal-free-queries}).
With coordinate encoding $\gamma$ and input
projection $\phi_{\mathrm{in}}$, queries and support features are
\begin{equation}
    \mathbf u_i=\phi_{\mathrm{in}}([\gamma(\mathbf c_i),\mathbf0,s_i^{\mathcal C}]),
    \qquad
    \mathbf h_j=\phi_{\mathrm{in}}([\gamma(\mathbf x_j),\mathbf n_j,s_j]).
    \label{eq:main-query}
\end{equation}
Encoder cross-attention uses $\mathbf u_i$ as queries and $\mathbf h_j$
as keys and values. The sharp-edge label is retained: normal-free does
not mean identical feature encodings in the two stages. Both stages
construct queries without a supplied normal, while the encoder still
observes orientation through its support. The locality comparison tests
how this change affects information collection.

\vspace{-4.8pt}
\paragraph{Profiling normal-free queries.}
Figure~\ref{fig:component-profiles}(c) compares encoder local selection,
the fraction of queries that select a neighboring input point, between
the complete models. Table~\ref{tab:diagnostic_summary} traces this change
and local attention mass $P_{\mathrm{enc},k}$ across cumulative variants.
Table~\ref{tab:query_normal}
repeats the perturbations from Section~\ref{sec:remaining-gaps}.
Section~\ref{sec:mechanism-analysis} connects these attention changes to
reconstruction quality.

\vspace{-4.8pt}
\subsection{Distance-Biased Retrieval}
\label{sec:decoder}

\vspace{-4.8pt}
Spatially repeated responses motivate an explicit spatial reference for
retrieval (Figure~\ref{fig:method-distance-bias}). An anchor head predicts
$\widehat{\mathbf a}_i$ for shape
token $i$, supervised against its FPS center by a regression loss
$\mathcal L_{\mathrm{anchor}}$. For content logit $s_{tn}^{h}$ in head
$h$, decoder attention is
\begin{equation}
    A_{tn}^{h}=\operatorname{softmax}_{n}(s_{tn}^{h}+b_{tn}),\qquad
    b_{tn}=\begin{cases}
        0,&n\leq K,\\
        -\dfrac{\beta_\tau}{2\sigma(M)^2}
        \|\mathbf q_t-\operatorname{sg}[\widehat{\mathbf a}_{n-K}]\|_2^2,
        &n>K.
    \end{cases}
    \label{eq:main-bias}
\end{equation}
Here $\operatorname{sg}$ stops the gradient through anchors in the prior;
anchor regression still trains them. Nearby anchors receive a smaller
penalty, while sufficiently strong content evidence can override the
preference. The softmax spans bank and shape tokens together. The
strength $\beta_\tau$ is warmed up, and the bandwidth $\sigma(M)$ can
adapt to the token budget. Decoding uses predicted anchors and therefore
does not require the original FPS coordinates. Attention remains dense;
its measured overhead is included in the efficiency comparison.

\vspace{-4.8pt}
\paragraph{Profiling distance-biased retrieval.}
Figure~\ref{fig:component-profiles}(b) compares response correlation
$C_{\mathrm{rep}}$ for VecSet and \method.
Table~\ref{tab:attention_profile} reports read distances and spreads
for the cumulative variants, using the original FPS centers as reference.
A separate bias toggle keeps the latent code and predicted anchors fixed.
The bias changes attention but need not remove repetition from raw logits.

\vspace{-4.8pt}
\subsection{Surface and Normal-Direction Supervision}
\label{sec:normal-supervision}

\vspace{-4.8pt}
Inspired by screened Poisson reconstruction~\citep{kazhdan2013screened},
we supervise both the zero crossing and the field's variation along an
observed outward normal (Figure~\ref{fig:method-normal-supervision}).
In the finite-difference formulation, the same
latent sample is queried at $\mathbf x$ and $\mathbf x\pm\varepsilon\mathbf n$:
\begin{equation}
    d^{\pm}=\frac{\hat f(\mathbf x\pm\varepsilon\mathbf n)-\hat f(\mathbf x)}
                         {\pm\varepsilon r},\qquad
    \mathcal L_{\mathrm{normal}}=
    \mathbb E\!\left[\lambda_{\mathrm{surf}}\rho(\hat f(\mathbf x))
    +\frac{\lambda_\tau}{2}\sum_{a\in\{+,-\}}\rho(d^a-1)\right].
    \label{eq:main-normal}
\end{equation}
The robust penalty $\rho$ is Smooth L1 and $\varepsilon r<1$ keeps
probes in the untruncated band. This constrains the normal direction,
not every tangential derivative; full field-normal accuracy is evaluated
independently. Appendix~\ref{sec:detail-normal-supervision} details the
three-point sampling and loss reduction.

\begin{figure}[tbp]
\centering
\includegraphics[width=.92\textwidth]{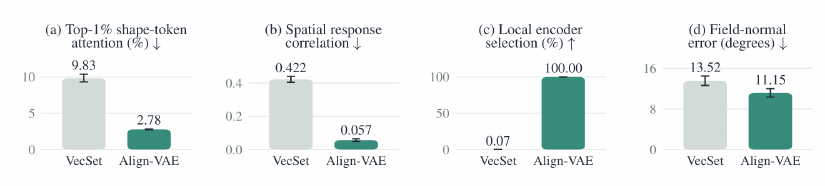}
\vspace{-4pt}
\caption{VecSet (gray) and \method (teal): 8192 shape latents, 64 matched
shapes. Bars show means and 95\% shape-bootstrap intervals.
Panel (a) conditions mass on shape tokens.}
\label{fig:component-profiles}
\vspace{-9pt}
\end{figure}

\vspace{-4.8pt}
\paragraph{Profiling normal supervision.}
Figure~\ref{fig:component-profiles}(d) compares field-normal angular error
between standard VecSet and the complete \method model.
The angular error $E_{\mathrm{angle}}$ measures agreement between the
predicted field gradient and the observed surface normal.

\vspace{-4.8pt}
\paragraph{Training objective.}
\label{sec:training-objective}
We optimize the combined objective:
\begin{equation}
    \mathcal L=\mathcal L_{\mathrm{field}}
    +\lambda_{\mathrm{KL}}\mathcal L_{\mathrm{KL}}
    +\lambda_{\mathrm{anchor}}\mathcal L_{\mathrm{anchor}}
    +\mathcal L_{\mathrm{normal}}.
    \label{eq:main-objective}
\end{equation}
There is no attention-balancing loss in this four-component method.
The cumulative ablation adds the bank (LB), distance bias and anchor
regression (DB), normal-free queries (PQ), and normal supervision (NS),
in that order.

\section{Experiments}
\label{sec:experiments}

\vspace{-4.8pt}
Section~\ref{sec:alignment-analysis} shows that spatially seeded tokens
can collect and retrieve information far from their intended neighborhoods.
We now ask whether aligning these operations improves reconstruction,
which components contribute to the gains, and how their attention changes.
Chamfer distance (CD) measures surface error, while F1 measures agreement
within a distance threshold; lower CD and higher F1 are better.
Appendix~\ref{sec:reconstruction-metrics} gives their mathematical definitions.

\vspace{-4.8pt}
\subsection{Reconstruction Quality and Compactness}

\vspace{-4.8pt}
\paragraph{Better reconstruction from the same code budget.}
Table~\ref{tab:main_recon} compares \method with standard VecSet VAE,
LATTICE, and released 3D autoencoders. At 4096 tokens, \method lowers
CD relative to VecSet by 18.2\% on Objaverse~\citep{deitke2023objaverse},
22.0\% on ABO~\citep{collins2022abo}, and
16.4\% on in-the-wild shapes. The gain reaches 26.6\% on ABO at
8192 tokens. F1@.001 also improves across all three datasets, indicating
closer surface agreement. At 8192 tokens, \method leads every reported
metric in Table~\ref{tab:main_recon}.

\vspace{-4.8pt}
\paragraph{Reconstruction across Dora Bench levels.}
Table~\ref{tab:dora_bench} extends the comparison to the four levels of
Dora Bench~\citep{chen2025dora}. At 8192 tokens, \method reduces CD by
48.1\% overall relative to VecSet, from 4.804 to 2.495, while F1@.001
increases from 56.87\% to 60.22\%. The largest CD improvement appears
on Level 4, where the error falls from 11.883 to 2.884.
\method achieves the lowest mean CD on all four levels, while VecSet
retains the highest F1@.001 on Level 3.
At 4096 tokens, overall CD falls from 3.402 to 2.804, indicating that
geometric alignment benefits reconstruction at both latent budgets.

\begin{figure}[t]
\centering
\includegraphics[width=\textwidth]{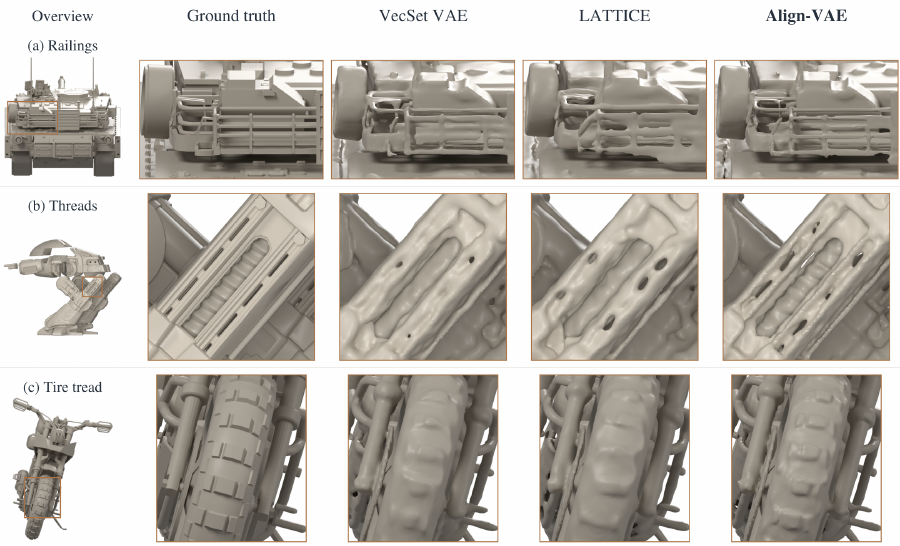}
\vspace{-4pt}
\caption{Local geometry at 8192 tokens on selected shapes.
Ground-truth overviews mark (a) rear railings,
(b) screw threads, and (c) tire tread.
Each row uses the same camera, material, lighting, and crop across methods.
Details are rendered directly at higher resolution.}
\label{fig:qualitative}
\vspace{-15pt}
\end{figure}

\vspace{-4.8pt}
\paragraph{Preserving local structure.}
Figure~\ref{fig:qualitative} compares \method with VecSet and LATTICE
at 8192 tokens on selected shapes. The enlarged views show better
separated railing bars, clearer screw threads, and raised tire-tread
blocks. These local differences are difficult to judge from the overall
silhouette alone.
These improvements need not reduce whole-shape reconstruction error.

\begin{table}[t]
\centering
\caption{Median reconstruction scores on Objaverse, ABO, and in-the-wild shapes.
Lower CD ($\times10^3$) and higher F1 (\%) are better. Throughout the main
text, \textbf{bold} and \underline{underlined} scores mark the best and
second-best per column. $\dagger$ marks our reproductions of VecSet and LATTICE.}
\label{tab:main_recon}
\small
\setlength{\tabcolsep}{4.2pt}
\begin{tabular}{@{}llccccccccc@{}}
\toprule
\groupedtablehead{3}{Method} & \groupedtablehead{3}{Tokens}
 & \multicolumn{3}{c}{Objaverse}
 & \multicolumn{3}{c}{ABO} & \multicolumn{3}{c}{In-the-wild} \\
\cmidrule(lr){3-5}\cmidrule(lr){6-8}\cmidrule(lr){9-11}
 & & \multirow[c]{2}{*}{CD\,$\downarrow$} & F1@ & F1@
   & \multirow[c]{2}{*}{CD\,$\downarrow$} & F1@ & F1@
   & \multirow[c]{2}{*}{CD\,$\downarrow$} & F1@ & F1@ \\
 & & & 0.01\,$\uparrow$ & 0.001\,$\uparrow$ & & 0.01\,$\uparrow$ & 0.001\,$\uparrow$
   & & 0.01\,$\uparrow$ & 0.001\,$\uparrow$ \\
\midrule
Dora & 4096 & 5.208 & 98.90 & 21.15 & 5.052 & 98.49 & 22.31 & 4.317 & \underline{99.97} & 24.03 \\
HY3D-2.1 & 4096 & 6.593 & 97.72 & 24.56 & 6.295 & 98.28 & 25.01 & 11.947 & 94.92 & 25.52 \\
VecSet VAE$^\dagger$ & 4096 & 4.797 & 98.97 & 28.93 & 5.143 & 98.74 & 26.32 & 3.017 & 99.83 & 44.33 \\
VecSet VAE$^\dagger$ & 8192 & \underline{3.773} & \underline{99.62} & 34.57 & \underline{3.927} & 99.54 & \underline{34.53} & \underline{2.479} & 99.95 & \underline{52.91} \\
LATTICE$^\dagger$ & 4096 & 5.529 & 98.11 & 26.20 & 5.490 & 97.83 & 24.24 & 3.773 & 99.68 & 34.86 \\
LATTICE$^\dagger$ & 8192 & 4.587 & 99.17 & 29.12 & 4.649 & 98.92 & 28.87 & 3.122 & 99.89 & 42.39 \\
\midrule
\method & 4096 & 3.923 & 99.57 & \underline{34.78} & 4.011 & \underline{99.68} & 31.53 & 2.523 & 99.96 & 50.60 \\
\method & 8192 & \textbf{3.079} & \textbf{99.88} & \textbf{42.35} & \textbf{2.884} & \textbf{99.90} & \textbf{41.20} & \textbf{2.238} & \textbf{99.99} & \textbf{54.68} \\

\bottomrule
\end{tabular}
\vspace{-15pt}
\end{table}

\begin{table}[t]
\centering
\caption{Mean reconstruction scores on Dora Bench. CD is scaled by $10^3$;
F1 denotes F1@0.001 in percent. Overall averages valid objects across all levels.
Method notation follows Table~\ref{tab:main_recon}.}
\label{tab:dora_bench}
\footnotesize
\setlength{\tabcolsep}{3pt}
\begin{tabular}{@{}llrrrrrrrrrr@{}}
\toprule
\groupedtablehead{2}{Method} & \groupedtablehead{2}{Tokens}
 & \multicolumn{2}{c}{Level 1}
 & \multicolumn{2}{c}{Level 2} & \multicolumn{2}{c}{Level 3}
 & \multicolumn{2}{c}{Level 4} & \multicolumn{2}{c}{Overall} \\
\cmidrule(lr){3-4}\cmidrule(lr){5-6}\cmidrule(lr){7-8}
\cmidrule(lr){9-10}\cmidrule(lr){11-12}
 & & CD\,$\downarrow$ & F1\,$\uparrow$
 & CD\,$\downarrow$ & F1\,$\uparrow$
 & CD\,$\downarrow$ & F1\,$\uparrow$
 & CD\,$\downarrow$ & F1\,$\uparrow$
 & CD\,$\downarrow$ & F1\,$\uparrow$ \\
\midrule
Dora & 4096 & 3.927 & 13.57 & 4.089 & 21.41 & 3.932 & 30.19 & 4.083 & 26.25 & 4.008 & 22.85 \\
HY3D-2.1 & 4096 & 99.117 & 11.78 & 66.173 & 17.04 & 58.317 & 15.15 & 13.855 & 24.43 & 59.365 & 17.10 \\
VecSet VAE$^\dagger$ & 4096 & 2.683 & 69.97 & 2.132 & 62.24 & 3.030 & 49.40 & 5.761 & 33.77 & 3.402 & 53.84 \\
VecSet VAE$^\dagger$ & 8192 & \underline{2.547} & \underline{74.21} & \underline{2.053} & \underline{64.00} & \underline{2.732} & \textbf{52.65} & 11.883 & 36.61 & 4.804 & \underline{56.87} \\
LATTICE$^\dagger$ & 4096 & 2.990 & 64.18 & 2.937 & 55.14 & 3.557 & 46.29 & 4.792 & 29.23 & 3.569 & 48.71 \\
LATTICE$^\dagger$ & 8192 & 3.033 & 65.97 & 2.658 & 56.27 & 3.320 & 49.22 & 4.086 & 35.73 & 3.278 & 51.57 \\
\midrule
\method & 4096 & 2.753 & 71.99 & 2.144 & 61.75 & 3.016 & 49.31 & \underline{3.305} & \underline{40.40} & \underline{2.804} & 55.86 \\
\method & 8192 & \textbf{2.445} & \textbf{79.87} & \textbf{1.943} & \textbf{65.21} & \textbf{2.709} & \underline{52.01} & \textbf{2.884} & \textbf{43.79} & \textbf{2.495} & \textbf{60.22} \\

\bottomrule
\end{tabular}
\vspace{-9pt}
\end{table}

\vspace{-4.8pt}
\paragraph{Storage and cost.}
At 4096 tokens, both models store 512~KiB per shape. The shared bank
increases the decoder sequence from $M$ to $M+K$ tokens while preserving
the $M C_z$ stored latent values. Latent refinement takes 56.32 ms
for \method and 53.99 ms for VecSet (Table~\ref{tab:efficiency}).

\vspace{-4.8pt}
\subsection{Cumulative Ablation}
\label{sec:ablations}

\vspace{-4.8pt}
\hypertarget{ablation-definitions}{A0} is standard VecSet. A1 adds a \emph{learnable decoder bank} (LB)
for shared context; A2 adds \emph{distance-biased retrieval} (DB)
around predicted anchors; A3 uses \emph{normal-free position queries}
(PQ); and A4 adds \emph{normal supervision} (NS) through surface and
normal-direction constraints. Each variant retains the preceding changes
at 4096 tokens.

\vspace{-4.8pt}
Table~\ref{tab:ablation} shows a steady reduction in CD from 3.723 to
2.526 (32.2\%), while F1@.001 rises from 35.15\% to 49.12\%.
The decoder bank supplies the largest initial gain; distance bias and
normal-free queries further improve reconstruction. Sharp-point coverage
(S-F) also increases, showing that the gains extend to fine geometry.

\begin{table}[t]
\centering
\caption{Cumulative ablation at 4096 tokens; A0--A4 are defined
in Section~\ref{sec:ablations}.
CD is scaled by $10^3$; F1 and sharp-point coverage S-F are percentages;
normal error $E_{\mathrm{angle}}$ is in degrees.}
\label{tab:ablation}
\footnotesize
\renewcommand{\arraystretch}{1.25}
\setlength{\tabcolsep}{5pt}
\begin{tabular}{lcccccccc}
\toprule
\groupedtablehead{3}{Variant}
  & \multicolumn{4}{c}{Component}
  & \multicolumn{4}{c}{Metric} \\
\cmidrule(lr){2-5} \cmidrule(lr){6-9}
 & \multirow[c]{2}{*}{Bank} & Distance & Normal-free & Normal
 & \multirow[c]{2}{*}{CD\,$\downarrow$}
 & \multirow[c]{2}{*}{F1@.001\,$\uparrow$}
 & \multirow[c]{2}{*}{S-F\,$\uparrow$}
 & \multirow[c]{2}{*}{$E_{\mathrm{angle}}$\,$\downarrow$} \\
 & & bias & queries & supervision & & & & \\
\midrule
\ablationvariant{A0} & $\times$ & $\times$ & $\times$ & $\times$ & 3.723 & 35.15 & 97.30 & 12.83 \\
\ablationvariant{A1} & $\checkmark$ & $\times$ & $\times$ & $\times$ & 2.960 & 42.70 & 98.30 & 11.92 \\
\ablationvariant{A2} & $\checkmark$ & $\checkmark$ & $\times$ & $\times$ & 2.809 & 45.99 & 98.44 & \underline{11.61} \\
\ablationvariant{A3} & $\checkmark$ & $\checkmark$ & $\checkmark$ & $\times$ & \underline{2.723} & \underline{46.71} & \underline{99.11} & \textbf{10.82} \\
\ablationvariant{A4} & $\checkmark$ & $\checkmark$ & $\checkmark$ & $\checkmark$ & \textbf{2.526} & \textbf{49.12} & \textbf{99.50} & 14.48 \\

\bottomrule
\end{tabular}
\vspace{-6pt}
\end{table}

\vspace{-4.8pt}
\subsection{How Does Geometric Alignment Improve?}
\label{sec:mechanism-analysis}

\begin{table}[!tb]
\centering
\caption{Geometric behavior of A0--A4,
defined in Section~\ref{sec:ablations}.
Local selection measures how often an encoder query selects one of its
64 nearest inputs; local mass sums attention to those neighbors.
Lower response correlation and normal error indicate less spatial repetition
and better orientation. Bank mass describes attention to shared tokens
and has no preferred direction.}
\label{tab:diagnostic_summary}
\footnotesize
\renewcommand{\arraystretch}{1.25}
\setlength{\tabcolsep}{4pt}
\begin{tabular}{lccccc}
\toprule
\multirow[c]{2}{*}{Variant} & Local selection & Local mass & Bank mass
 & Response corr. & Normal error \\
 & $L_{\mathrm{enc}}$ (\%)\,$\uparrow$
 & $P_{\mathrm{enc},k}$ (\%)\,$\uparrow$
 & $G_{\mathrm{bank}}$ (\%)
 & $C_{\mathrm{rep}}$\,$\downarrow$
 & $E_{\mathrm{angle}}$ (deg)\,$\downarrow$ \\
\midrule
\ablationvariant{A0} & \underline{0.076} & 3.05 & 0.00 & 0.392 & 12.83 \\
\ablationvariant{A1} & 0.063 & 13.80 & 31.19 & 0.344 & 11.92 \\
\ablationvariant{A2} & 0.043 & 9.77 & 54.27 & 0.052 & \underline{11.61} \\
\ablationvariant{A3} & \textbf{100.000} & \textbf{53.82} & 53.17 & \underline{0.034} & \textbf{10.82} \\
\ablationvariant{A4} & \textbf{100.000} & \underline{46.73} & 48.69 & \textbf{0.031} & 14.48 \\

\bottomrule
\end{tabular}
\vspace{-12pt}
\end{table}

\vspace{-4.8pt}
Figure~\ref{fig:component-profiles} revisits the profiles from
Section~\ref{sec:alignment-analysis} at 8192 latents. Relative to VecSet,
\method reduces response correlation from 0.422~to~0.057
and top-1\% shape-token attention from 9.83\% to 2.78\%. Local encoder selection
rises from 0.073\% to nearly 100\%, while normal error falls from
$13.52^\circ$ to $11.15^\circ$. Table~\ref{tab:diagnostic_summary}
examines the component contributions through cumulative variants at
4096 tokens, using the same sequence of additions as the reconstruction
ablation in Table~\ref{tab:ablation}.

\vspace{-6.8pt}
\paragraph{Separating context from spatial retrieval.}
In A1, the learnable bank receives 31.19\% of surface-query attention,
providing a route for shared context. Adding distance bias in A2
reduces the mean read distance from 0.175 to 0.035 and response
correlation from 0.344 to 0.052. These changes address the displaced
and repeated retrieval patterns observed in the baseline while the
bank remains available for shared context.

\vspace{-6.8pt}
\paragraph{Collecting evidence near each encoder query.}
With normal-free queries, A3 selects a neighboring input point for every
encoder query, compared with 0.043\% in A2. The attention assigned to
nearby inputs also rises from 9.77\% to 53.82\%, so the change extends
beyond the single largest attention weight. Reconstruction remains
unchanged when query normals are zeroed, perturbed, or flipped
(Table~\ref{tab:query_normal}). Removing the unavailable query cue thus
aligns the encoder's information collection with the decoder's inputs.

\vspace{-4.8pt}
\paragraph{Reconstruction and orientation are complementary.}
The combined A4 variant gives the best reconstruction scores, but its
field-normal error rises from $10.82^\circ$ to $14.48^\circ$ relative
to A3. Surface accuracy and gradient orientation therefore capture
different aspects of the learned field. This distinction explains why
we inspect geometry and attention alongside reconstruction scores;
Appendix~\ref{sec:extended-results} expands these comparisons.

\ifdefined\arxivpreprint
  \newpage
\fi
\vspace{-4.8pt}
\section{Conclusion}
\label{sec:conclusion}

\vspace{-4.8pt}
A token's spatial origin does not guarantee geometrically aligned
attention. \method combines a shared decoder workspace, normal-free
encoder queries, spatially biased retrieval, and field supervision while
preserving the stored code budget. The resulting representation improves
reconstruction and exhibits more local encoder selection and decoder
reads. Reconstruction is also insensitive to perturbing encoder query
normals. The shared bank keeps per-shape storage fixed but adds decoder
computation, so storage efficiency and inference cost must be evaluated
separately. The distinction between surface accuracy and field orientation
also shows the value of examining how information moves through a 3D
representation, beyond measuring its reconstruction error alone.
Future work should investigate how to balance surface reconstruction and
field orientation, particularly when normal supervision improves one but
not the other.

\ifdefined\arxivpreprint\else
  \clearpage
\fi
\subsection*{AI Use Statement}

We used generative AI tools to polish and rewrite the manuscript text and
to help search the literature and format references. We did not use these
tools to formulate hypotheses, design experiments, or interpret results. The authors reviewed all AI-assisted text and checked
every reference against its original source. We take responsibility for
the final content of this work, including text produced with the aid of
generative AI.

\bibliography{iclr2027_conference}

@inproceedings{kingma2013auto,
  title     = {Auto-Encoding Variational Bayes},
  author    = {Kingma, Diederik P. and Welling, Max},
  booktitle = {International Conference on Learning Representations},
  year      = {2014},
  url       = {https://arxiv.org/abs/1312.6114}
}

@inproceedings{collins2022abo,
  title     = {{ABO}: Dataset and Benchmarks for Real-World {3D} Object Understanding},
  author    = {Collins, Jasmine and Goel, Shubham and Deng, Kenan and Luthra, Achleshwar and Xu, Leon and Gundogdu, Erhan and Zhang, Xi and Vicente, Tomas F. Yago and Dideriksen, Thomas and Arora, Himanshu and Guillaumin, Matthieu and Malik, Jitendra},
  booktitle = {Proceedings of the IEEE/CVF Conference on Computer Vision and Pattern Recognition},
  pages     = {21126--21136},
  year      = {2022}
}

@inproceedings{deitke2023objaverse,
  title     = {{Objaverse}: A Universe of Annotated {3D} Objects},
  author    = {Deitke, Matt and Schwenk, Dustin and Salvador, Jordi and Weihs, Luca and Michel, Oscar and VanderBilt, Eli and Schmidt, Ludwig and Ehsani, Kiana and Kembhavi, Aniruddha and Farhadi, Ali},
  booktitle = {Proceedings of the IEEE/CVF Conference on Computer Vision and Pattern Recognition},
  pages     = {13142--13153},
  year      = {2023}
}

@inproceedings{zhang20223dilg,
  title   = {{3DILG}: Irregular Latent Grids for {3D} Generative Modeling},
  author  = {Zhang, Biao and Nie{\ss}ner, Matthias and Wonka, Peter},
  booktitle = {Advances in Neural Information Processing Systems},
  volume  = {35},
  pages   = {21871--21885},
  year    = {2022}
}

@article{zhang20233dshape2vecset,
  title     = {{3DShape2VecSet}: A {3D} Shape Representation for Neural Fields and Generative Diffusion Models},
  author    = {Zhang, Biao and Tang, Jiapeng and Nie{\ss}ner, Matthias and Wonka, Peter},
  journal   = {ACM Transactions on Graphics},
  volume    = {42},
  number    = {4},
  pages     = {1--16},
  year      = {2023},
  publisher = {ACM},
  doi       = {10.1145/3592442}
}

@inproceedings{zhao2023michelangelo,
  title     = {Michelangelo: Conditional {3D} Shape Generation Based on Shape-Image-Text Aligned Latent Representation},
  author    = {Zhao, Zibo and Liu, Wen and Chen, Xin and Zeng, Xianfang and Wang, Rui and Cheng, Pei and Fu, Bin and Chen, Tao and Yu, Gang and Gao, Shenghua},
  booktitle = {Advances in Neural Information Processing Systems},
  volume    = {36},
  pages     = {73969--73982},
  year      = {2023}
}

@inproceedings{ren2024xcube,
  title     = {{XCube}: Large-Scale {3D} Generative Modeling Using Sparse Voxel Hierarchies},
  author    = {Ren, Xuanchi and Huang, Jiahui and Zeng, Xiaohui and Museth, Ken and Fidler, Sanja and Williams, Francis},
  booktitle = {Proceedings of the IEEE/CVF Conference on Computer Vision and Pattern Recognition},
  pages     = {4209--4219},
  year      = {2024}
}

@article{zhang2024clay,
  title     = {{CLAY}: A Controllable Large-Scale Generative Model for Creating High-Quality {3D} Assets},
  author    = {Zhang, Longwen and Wang, Ziyu and Zhang, Qixuan and Qiu, Qiwei and Pang, Anqi and Jiang, Haoran and Yang, Wei and Xu, Lan and Yu, Jingyi},
  journal   = {ACM Transactions on Graphics},
  volume    = {43},
  number    = {4},
  pages     = {1--20},
  year      = {2024},
  publisher = {ACM},
  doi       = {10.1145/3658146}
}

@inproceedings{xiang2025structured,
  title     = {Structured {3D} Latents for Scalable and Versatile {3D} Generation},
  author    = {Xiang, Jianfeng and Lv, Zelong and Xu, Sicheng and Deng, Yu and Wang, Ruicheng and Zhang, Bowen and Chen, Dong and Tong, Xin and Yang, Jiaolong},
  booktitle = {Proceedings of the IEEE/CVF Conference on Computer Vision and Pattern Recognition},
  pages     = {21469--21480},
  year      = {2025}
}

@inproceedings{wu2025direct3d,
  title     = {{Direct3D-S2}: Gigascale {3D} Generation Made Easy with Spatial Sparse Attention},
  author    = {Wu, Shuang and Lin, Youtian and Zhang, Feihu and Zeng, Yifei and Yang, Yikang and Bao, Yajie and Qian, Jiachen and Zhu, Siyu and Cao, Xun and Torr, Philip and Yao, Yao},
  booktitle = {Advances in Neural Information Processing Systems},
  volume    = {38},
  pages     = {170778--170804},
  year      = {2025},
  doi       = {10.52202/085713-5688}
}

@article{li2025sparc3d,
  title   = {{Sparc3D}: Sparse Representation and Construction for High-Resolution {3D} Shapes Modeling},
  author  = {Li, Zhihao and Wang, Yufei and Zheng, Heliang and Luo, Yihao and Wen, Bihan},
  journal = {arXiv preprint arXiv:2505.14521},
  year    = {2025},
  url     = {https://arxiv.org/abs/2505.14521}
}

@inproceedings{chen2025dora,
  title     = {Dora: Sampling and Benchmarking for {3D} Shape Variational Auto-Encoders},
  author    = {Chen, Rui and Zhang, Jianfeng and Liang, Yixun and Luo, Guan and Li, Weiyu and Liu, Jiarui and Li, Xiu and Long, Xiaoxiao and Feng, Jiashi and Tan, Ping},
  booktitle = {Proceedings of the IEEE/CVF Conference on Computer Vision and Pattern Recognition},
  pages     = {16251--16261},
  year      = {2025}
}

@inproceedings{lai2025lattice,
  title     = {{LATTICE}: Democratize High-Fidelity {3D} Generation at Scale},
  author    = {Lai, Zeqiang and Zhao, Yunfei and Zhao, Zibo and Liu, Haolin and Lin, Qingxiang and Huang, Jingwei and Guo, Chunchao and Yue, Xiangyu},
  booktitle = {Proceedings of the IEEE/CVF Conference on Computer Vision and Pattern Recognition},
  pages     = {19982--19992},
  year      = {2026},
  url       = {https://arxiv.org/abs/2512.03052}
}

@article{li2025triposg,
  title     = {{TripoSG}: High-Fidelity {3D} Shape Synthesis Using Large-Scale Rectified Flow Models},
  author    = {Li, Yangguang and Zou, Zi-Xin and Liu, Zexiang and Wang, Dehu and Liang, Yuan and Yu, Zhipeng and Liu, Xingchao and Guo, Yuan-Chen and Liang, Ding and Ouyang, Wanli and Cao, Yan-Pei},
  journal   = {IEEE Transactions on Pattern Analysis and Machine Intelligence},
  year      = {2025},
  publisher = {IEEE},
  pages     = {1--18},
  doi       = {10.1109/TPAMI.2025.3633512}
}

@article{hunyuan3d2025hunyuan3d,
  title   = {{Hunyuan3D 2.1}: From Images to High-Fidelity {3D} Assets with Production-Ready {PBR} Material},
  author  = {{Team Hunyuan3D} and Yang, Shuhui and Yang, Mingxin and Feng, Yifei and Huang, Xin and Zhang, Sheng and He, Zebin and Luo, Di and Liu, Haolin and Zhao, Yunfei and Lin, Qingxiang and Lai, Zeqiang and Yang, Xianghui and Shi, Huiwen and Zhao, Zibo and Zhang, Bowen and Yan, Hongyu and Wang, Lifu and Liu, Sicong and Zhang, Jihong and Chen, Meng and Dong, Liang and Jia, Yiwen and Cai, Yulin and Yu, Jiaao and Tang, Yixuan and Guo, Dongyuan and Yu, Junlin and Zhang, Hao and Ye, Zheng and He, Peng and Wu, Runzhou and Wei, Shida and Zhang, Chao and Tan, Yonghao and Sun, Yifu and Niu, Lin and Huang, Shirui and Zheng, Bojian and Liu, Shu and Chen, Shilin and Yuan, Xiang and Yang, Xiaofeng and Liu, Kai and Zhu, Jianchen and Chen, Peng and Liu, Tian and Wang, Di and Liu, Yuhong and {Linus} and Jiang, Jie and Huang, Jingwei and Guo, Chunchao},
  journal = {arXiv preprint arXiv:2506.15442},
  year    = {2025},
  url     = {https://arxiv.org/abs/2506.15442}
}

@article{hao2026msvsvae,
  title   = {{MSVS-VAE}: Multi-Scale Anchored {VecSet} for High-Fidelity {3D} Reconstruction},
  author  = {Hao, Dehao and Zhang, Kaiyi and Jia, Tanghui and Gao, Xiangjun and Yan, Dongyu and Chen, Weikai and Hu, Zeyu and Zhu, Lingting and Yin, Yingda and Zhang, Runze and Yuan, Li and Wang, Xin and Quan, Long},
  journal = {arXiv preprint arXiv:2607.24436},
  year    = {2026},
  url     = {https://arxiv.org/abs/2607.24436}
}

@inproceedings{kazhdan2006poisson,
  title     = {{Poisson} Surface Reconstruction},
  author    = {Kazhdan, Michael and Bolitho, Matthew and Hoppe, Hugues},
  booktitle = {Symposium on Geometry Processing},
  pages     = {61--70},
  year      = {2006},
  publisher = {The Eurographics Association},
  doi       = {10.2312/SGP/SGP06/061-070}
}

@article{kazhdan2013screened,
  title     = {Screened {Poisson} Surface Reconstruction},
  author    = {Kazhdan, Michael and Hoppe, Hugues},
  journal   = {ACM Transactions on Graphics},
  volume    = {32},
  number    = {3},
  pages     = {29:1--29:13},
  year      = {2013},
  doi       = {10.1145/2487228.2487237},
  url       = {https://hhoppe.com/proj/screenedpoisson/}
}

@inproceedings{darcet2024registers,
  title = {Vision Transformers Need Registers},
  author = {Darcet, Timoth{\'e}e and Oquab, Maxime and Mairal, Julien and Bojanowski, Piotr},
  booktitle = {International Conference on Learning Representations},
  year = {2024},
  url = {https://openreview.net/forum?id=2dnO3LLiJ1}
}

@inproceedings{xiao2024efficient,
  title={Efficient streaming language models with attention sinks},
  author={Xiao, Guangxuan and Tian, Yuandong and Chen, Beidi and Han, Song and Lewis, Mike},
  booktitle={International Conference on Learning Representations},
  volume={2024},
  pages={21875--21895},
  year={2024},
  url={https://openreview.net/forum?id=NG7sS51zVF}
}
\bibliographystyle{iclr2027_conference}

\clearpage
\appendix
\section{Implementation and Training Details}
\label{sec:method-details}

This appendix follows the four components in Figure~\ref{fig:method},
then collects loss reductions and training settings. Architectural
operations already defined in the main text are referenced directly.

\subsection{Learnable Decoder Bank}
\label{sec:detail-learnable-latents}

Equation~\ref{eq:main-bank} prepends $K$ learned tokens of width $d$ to
the projected $M$-token variational code. Joint decoder self-attention
makes the bank shape-dependent, and spatial-query cross-attention reads
all $K+M$ refined tokens. The bank has no direct access to the input
surface and enters only after the variational bottleneck. Its $Kd$
shared parameters add decoder capacity without changing the stored
per-shape code or the KL penalty.

Bank tokens have no assigned coordinates and receive no distance
penalty. They provide an unanchored route for context, while the shape
tokens support spatially addressed retrieval. Both populations compete
in the same softmax, with no prescribed attention quota for the bank.

\subsection{Distance-Biased Retrieval}
\label{sec:detail-decoder}

An anchor head predicts
$\widehat{\mathbf a}_i=g_a(\widetilde{\mathbf z}_{K+i})$ from refined
shape-token features. Training targets the FPS centers $\mathbf c_i$,
with the loss averaged over all token coordinates:
\begin{equation*}
    \mathcal L_{\mathrm{anchor}}=
    \frac{1}{3M}\sum_{i=1}^{M}\sum_{c=1}^{3}
       h_{\delta_a}(\widehat a_{ic}-c_{ic}).
\end{equation*}
Here $h_{\delta_a}$ is Huber loss: quadratic $u^2/2$ for residuals
$|u|\leq\delta_a$, and linear $\delta_a(|u|-\delta_a/2)$ otherwise.
The cumulative variants use squared error in place of Huber loss;
Table~\ref{tab:implementation-settings} records the applicable settings.
Anchors are detached only in the distance prior of
equation~\ref{eq:main-bias}; regression still trains the anchor head
and its input features. Decoding therefore needs no original FPS points.

The prior is shared across attention heads. After a content-only warm-up,
its strength $\beta_\tau$ rises linearly to $\beta$. The bandwidth adapts
to the number of shape tokens:
\begin{equation*}
    \sigma(M)=\sigma_{\mathrm{ref}}\sqrt{M_{\mathrm{ref}}/M}.
\end{equation*}
The reference bandwidth $\sigma_{\mathrm{ref}}$ is specified at token
count $M_{\mathrm{ref}}$. This scaling follows sample spacing on a
surface; it does not enforce a fixed number of attended neighbors.
The biased attention aggregates values inside the residual
cross-attention block, followed by normalization and a linear scalar
head that predicts the field value $\hat f(\mathbf q;\bm Z)$.

\subsection{Normal-Free Encoder Queries}
\label{sec:detail-position-only-queries}

Equation~\ref{eq:main-query} masks only the query normal channels.
Support keys and values retain observed normals, and both branches
retain their sharp-edge labels. Masking preserves the feature width
and introduces no parameters. The encoder and decoder thus construct
queries without supplied normals, although their projections and
remaining features differ.

Encoder cross-attention uses scaled dot products, query/key
normalization, and a softmax over support points. The residual block
includes value aggregation and a feed-forward sublayer. Encoder
self-attention then refines the $M$ tokens, and a projection predicts
the mean $\boldsymbol\mu$ and log variance of a diagonal Gaussian
posterior~\citep{kingma2013auto}. We sample $\bm Z$ from this posterior
during training; the decoder bank is added afterward.

\subsection{Surface and Normal-Direction Supervision}
\label{sec:detail-normal-supervision}

For each of $N_s$ surface samples, normalize the outward normal
$\mathbf n_k$ and evaluate the field at $\mathbf x_k$ and
$\mathbf x_k\pm\varepsilon\mathbf n_k$, as in Figure~\ref{fig:method}.
All three queries reuse the same sampled latent set and refined decoder
features. The two slopes $d_k^+$ and $d_k^-$ use the definition in
equation~\ref{eq:main-normal}, with the TSDF scale $r$ in their denominator.
Both approximate the field derivative along the normal divided by $r$,
whose target is one at a smooth surface point.

The sample reduction of this three-point constraint is
\begin{equation*}
    \mathcal L_{\mathrm{normal}}=
    \frac{\lambda_{\mathrm{surf}}}{N_s}\sum_{k=1}^{N_s}
       \rho(\hat f(\mathbf x_k))
    +\frac{\lambda_\tau}{2N_s}\sum_{k=1}^{N_s}
       \bigl[\rho(d_k^+-1)+\rho(d_k^--1)\bigr].
\end{equation*}
The Smooth L1 penalty $\rho$ is quadratic $u^2/2$ for $|u|<1$ and
linear $|u|-1/2$ otherwise. The surface weight is constant; the slope
weight $\lambda_\tau$ rises linearly from zero to
$\lambda_{\mathrm{normal}}$ during warm-up. The condition
$\varepsilon r<1$ keeps the target probes within the untruncated band;
the offset must also be small relative to local geometric variation.
The prior is evaluated at each displaced query, so the differences
include its effect on the predicted field. These probes constrain
variation along the normal. The six coordinate-axis probes in
Appendix~\ref{sec:profiling-guide} independently evaluate the complete
gradient direction.

\subsection{Loss Reductions and Training Settings}
\label{sec:detail-training-objective}
\label{sec:implementation-details}

The field scale $r$ converts normalized signed distance into the TSDF
target $y_t$ defined in Section~\ref{sec:preliminaries}. We average
squared errors separately over the volumetric query set $\mathcal V$
and near-surface set $\mathcal N$, of sizes $N_{\mathrm{vol}}$ and
$N_{\mathrm{near}}$:
\begin{equation*}
    \mathcal L_{\mathrm{field}}=
    \frac{1}{N_{\mathrm{vol}}}\sum_{t\in\mathcal V}
       (\hat f(\mathbf q_t)-y_t)^2+
    \frac{\lambda_{\mathrm{near}}}{N_{\mathrm{near}}}
       \sum_{t\in\mathcal N}(\hat f(\mathbf q_t)-y_t)^2.
\end{equation*}
The standard diagonal-Gaussian KL penalty against a unit Gaussian is
averaged over the $M$ posterior tokens and $C_z$ channels, excluding
the shared bank. The combined objective in
equation~\ref{eq:main-objective} is then averaged over shapes.
Anchor regression trains the retrieval geometry, while normal
supervision acts on the field; neither balances attention mass.

Table~\ref{tab:implementation-settings} collects settings from the frozen
training snapshots for the in-the-wild experiments. A0--A4 denote the cumulative
variants in Appendix~\ref{sec:extended-results}; ``Full'' denotes the
full-scale \method reproduction evaluated at 4096 and 8192 tokens.
A0--A3 use 100k-step checkpoints and A4 uses 110k steps, so this series
is not matched in training duration. Each parameter row states its
experimental scope.

\begin{table}[htbp]
\centering
\caption{Training settings for the in-the-wild experiments.
Offsets and bandwidths use normalized coordinates; schedule lengths
are optimizer steps. Settings are scoped to the listed experiments.}
\label{tab:implementation-settings}
\begin{tabular}{@{}p{.16\textwidth}p{.26\textwidth}p{.25\textwidth}p{.23\textwidth}@{}}
\toprule
Parameter & Meaning & Setting & Experiments \\
\midrule
$K$ & Decoder bank size & 256 & A1--A4, Full \\
$\sigma_{\mathrm{ref}}$ & Reference bandwidth & 0.097 & A2--A4, Full \\
$M_{\mathrm{ref}}$ & Reference token count & 4096; 8192 & A2--A4; Full \\
$\beta$ & Final prior strength & 1 & A2--A4, Full \\
Prior schedule & \raggedright Delay, then ramp & 10k, then 5k & A2--A4, Full \\
$\lambda_{\mathrm{anchor}}$ & Anchor loss weight & 0.1 & A2--A4, Full \\
Anchor loss & Coordinate penalty & \raggedright MSE; Huber ($\delta_a=0.04$) & A2--A4; Full \\
$N_{\mathrm{vol}}$, $N_{\mathrm{near}}$ & Queries per group & 4096; 8192 & A0--A4; Full \\
$\lambda_{\mathrm{near}}$ & Near-surface weight & 0.1 & A0--A4, Full \\
$\lambda_{\mathrm{KL}}$ & KL weight & $10^{-5}$ & A0--A4, Full \\
$r$ & Scale at $M\geq4096$ & 128 & A0--A4, Full \\
$N_s$, $\varepsilon$ & \raggedright Samples, offset & 2048, 0.004 & A4, Full \\
$\lambda_{\mathrm{surf}}$, $\lambda_{\mathrm{normal}}$ & Surface, slope weights & 1, 0.05 & A4, Full \\
Slope ramp & Linear warm-up & 10k & A4, Full \\
\bottomrule
\end{tabular}
\end{table}

\section{Evaluation and Profiling Protocols}

\subsection{Reconstruction Metrics}
\label{sec:reconstruction-metrics}

\paragraph{Chamfer distance (CD).}
Let $\mathcal M$ and $\widehat{\mathcal M}$ denote the ground-truth and
reconstructed meshes in normalized coordinates, with surface samples
$G$ and $P$ of sizes $N_G$ and $N_P$. The point-to-mesh distance is
$d(\mathbf x,\mathcal S)=\min_{\mathbf y\in\mathcal S}
\lVert\mathbf x-\mathbf y\rVert_2$; the closest point can lie inside
a triangle, on an edge, or at a vertex. CD sums both directional means:
\begin{equation*}
    \operatorname{CD}=
    \frac{1}{N_P}\sum_{\mathbf p\in P}d(\mathbf p,\mathcal M)
    +\frac{1}{N_G}\sum_{\mathbf g\in G}d(\mathbf g,\widehat{\mathcal M}).
\end{equation*}
Distances are unsquared, and the two terms are added without a factor
of $1/2$. CD therefore retains the units of the normalized geometry.

\paragraph{F1 score.}
At threshold $\delta$, precision $\operatorname{Prec}_{\delta}$ is the
fraction of predicted samples within distance $\delta$ of the true mesh;
recall $\operatorname{Rec}_{\delta}$ is the fraction of true samples
within that distance of the reconstructed mesh. Their harmonic mean is
\begin{equation*}
    \operatorname{F1}_{\delta}=
    \frac{2\operatorname{Prec}_{\delta}\operatorname{Rec}_{\delta}}
         {\operatorname{Prec}_{\delta}+\operatorname{Rec}_{\delta}}.
\end{equation*}
We set F1 to zero when both precision and recall are zero. F1@.01 and
F1@.001 use thresholds 0.01 and 0.001 in normalized coordinates;
a column labeled F uses 0.01. Each score is computed per shape before
aggregation across the evaluated objects.

\paragraph{Sharp-region distance and coverage.}
From a fixed ground-truth sharp set $G_{\mathrm{sharp}}$ shared by all
methods, we measure one-way distance and coverage:
\begin{align*}
    \operatorname{S\text{-}CD}
    &=\frac{1}{|G_{\mathrm{sharp}}|}\sum_{\mathbf g\in G_{\mathrm{sharp}}}
       d(\mathbf g,\widehat{\mathcal M}),\\
    \operatorname{S\text{-}F}_{\delta}
    &=\frac{1}{|G_{\mathrm{sharp}}|}\sum_{\mathbf g\in G_{\mathrm{sharp}}}
       \mathbf 1[d(\mathbf g,\widehat{\mathcal M})\leq\delta].
\end{align*}
S-F is the fraction of ground-truth sharp points within $\delta=0.01$
of the reconstruction, unless another threshold is stated. It is
one-way coverage and does not include a precision term.

\paragraph{Aggregation and units.}
Reconstruction tables report medians across shapes, except Dora Bench
(Table~\ref{tab:dora_bench}), which reports arithmetic means. CD and
S-CD are multiplied by $10^3$; F1 and S-F are percentages. Attention
and field-normal profiles report means across shapes, as defined below.

\subsection{Diagnostic Profiles}
\label{sec:diagnostics}

\subsubsection{Reading and Computing the Profiles}
\label{sec:profiling-guide}

\begin{figure}[p]
\centering
\includegraphics[width=\textwidth]{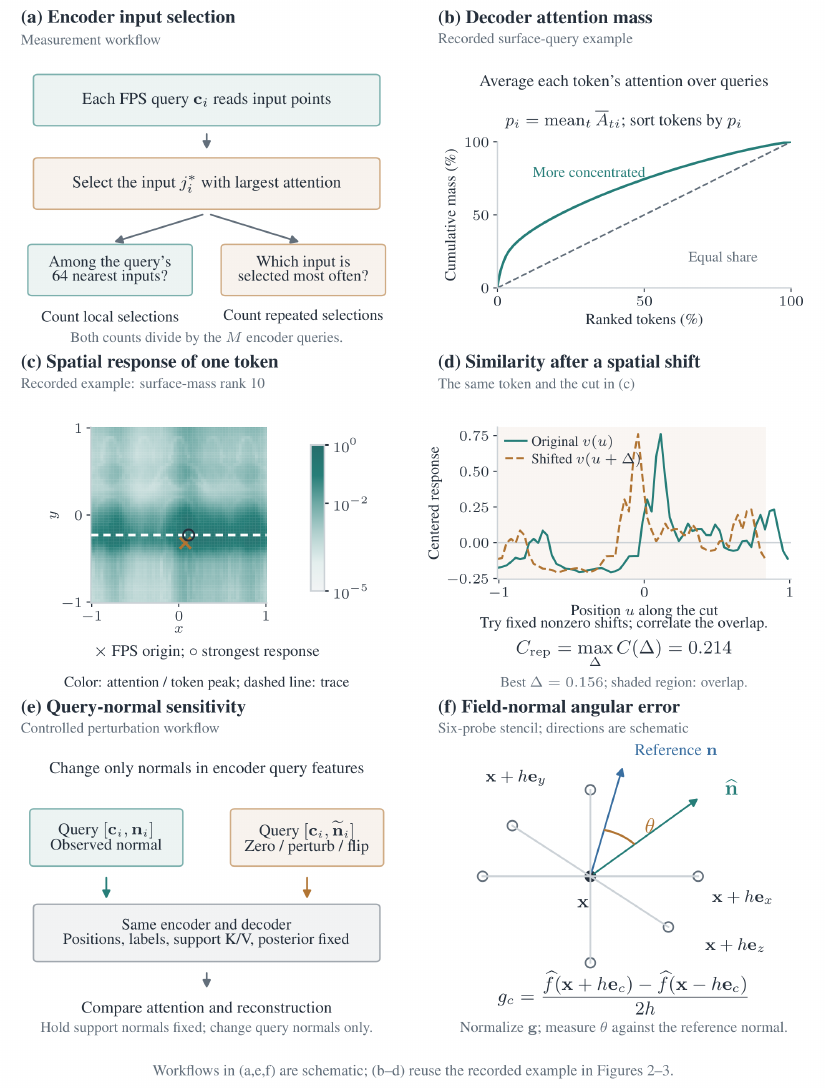}
\caption{A visual guide to the profiling measurements.
(a) Follow each encoder query's strongest attention link, then count local
or repeated selections.
(b) Average decoder attention over queries, rank tokens, and accumulate
their shares.
(c) Hold a token fixed and view its response over spatial locations.
(d) Shift a response curve and compare the overlapping values; retain
the largest correlation over the specified shifts.
(e) Change query normals while fixing the other encoder inputs.
(f) Estimate the field gradient with six offset queries and compare its
direction with the reference normal.
Panels (a,e,f) are schematics. Panels (b--d) reuse the first example in
Figures~\ref{fig:attention-diagnostics} and~\ref{fig:periodic-responses};
(c,d) use its surface-mass rank-10 token. Panel (d) reports the score
for this single trace.}
\label{fig:profiling-guide}
\end{figure}

Figure~\ref{fig:profiling-guide} links each diagnostic to its computation.
An encoder attention row corresponds to an FPS query and its columns
to input points; a decoder row corresponds to a field query and its
columns to latent tokens. Each softmax row sums to one. We average
weights over heads before computing the main attention statistics,
then compute each statistic per shape and average across shapes.
Figure~\ref{fig:component-profiles} compares complete models at 8192
shape tokens using the same inputs, FPS centers, and spatial probes.
Table~\ref{tab:profiling-metrics} provides a symbol and unit lookup.

\paragraph{(a) Local and repeated encoder selection.}
For each FPS query, select the input with the largest head-averaged
attention weight. Local selection $L_{\mathrm{enc}}$ is the fraction
of queries whose selected input belongs to their $k=64$ nearest
support points. The most-selected input share is the largest selection
count divided by the number of queries. These events may overlap:
the two bars in Figure~\ref{fig:encoder-locality-statistics} are not a partition.
The per-head control instead takes each head's argmax first and
averages the fraction of local selections across heads.

Local mass $P_{\mathrm{enc},k}$ sums the attention on the same 64
neighbors, then averages over queries. Comparing this mass with local
selection separates total local evidence from the strongest link.
The decoder bank contributes no encoder rows or support points.

\clearpage
\paragraph{(b) Decoder mass and concentration.}
Let $\overline A_{t,K+i}$ be the head-averaged decoder attention from
query $t$ to shape token $i$, after the $K$ bank columns. Summing over
queries and normalizing by total shape-token attention gives its share
$p_i$ and the normalized entropy $H_{\mathrm{shape}}$:
\begin{equation*}
    p_i=\frac{\sum_t\overline A_{t,K+i}}
               {\sum_t\sum_{j=1}^{M}\overline A_{t,K+j}},
    \qquad
    H_{\mathrm{shape}}=-\frac{1}{\log M}
       \sum_{i:p_i>0}p_i\log p_i.
\end{equation*}
Equal allocation gives entropy one; concentration on one token gives
zero. Sorting $p_i$ and accumulating it yields the mass curve. Its
first $\lceil0.01M\rceil$ entries sum to the top-1\% share $T_{1\%}$.
Ranks are computed independently for each model and probe population,
excluding the bank. Bank share $G_{\mathrm{bank}}$ is the sum over bank
columns, averaged over queries; it is zero when $K=0$. Conditioning on
shape tokens uses the denominator above, rather than averaging
separately normalized query rows. A zero denominator leaves the
conditional profile undefined.

\paragraph{(c) Spatial response, read distance, and spread.}
Holding a token fixed gives its attention response over query locations.
Maps divide each response by its own peak to show spatial structure;
color thus indicates relative response, not absolute attention mass.
The token's FPS center supplies a common reference for both models.
The conditional mean read distance is
\begin{equation*}
    R_{\mathrm{shape}}=
    \frac{\sum_t\sum_{i=1}^{M}\overline A_{t,K+i}
                    \lVert\mathbf q_t-\mathbf c_i\rVert_2}
         {\sum_t\sum_{i=1}^{M}\overline A_{t,K+i}}.
\end{equation*}
The denominator conditions on shape-token reads, so diverting attention
to the bank alone cannot reduce this distance. Restricting both sums
to the most-read 1\% of shape tokens gives $R_{\mathrm{top}}$.
Anchor error, the mean Euclidean distance between predicted anchors
and FPS centers, checks the predictor separately from these read distances.

For token $i$, normalize its response across queries as
$w_{ti}=\overline A_{t,K+i}/\sum_u\overline A_{u,K+i}$.
Its response centroid $\bar{\mathbf q}_i$ and root-mean-square spread
$S_i$ are
\begin{equation*}
    \bar{\mathbf q}_i=\sum_t w_{ti}\mathbf q_t,\qquad
    S_i=\left(\sum_t w_{ti}
             \lVert\mathbf q_t-\bar{\mathbf q}_i\rVert_2^2\right)^{1/2}.
\end{equation*}
The mean spread at mass ranks 1, 10, and the median is $S_{\mathrm{sel}}$;
$S_{\mathrm{top}}$ is the attention-mass-weighted mean of $S_i$ over
the top 1\%. Zero-mass tokens have undefined conditional statistics.
Table~\ref{tab:figure1-measured} includes uniform references that give
equal weight to each spatial query. A compact but displaced response
can have a large read distance; a broad response has large spread
around its own centroid.

\paragraph{(d) Similarity across spatial shifts.}
For a token's line response $a(u)$, subtract the mean of the entire line
to obtain $v(u)=a(u)-\overline a$. For shift $\ell$, let $I_\ell$
contain only samples that overlap with the shifted line, without
wrapping at the boundaries. The normalized correlation and its maximum
are
\begin{equation*}
    C(\ell)=\frac{\sum_{u\in I_\ell}v(u)v(u+\ell)}
       {\sqrt{\sum_{u\in I_\ell}v(u)^2
                    \sum_{u\in I_\ell}v(u+\ell)^2}},\qquad
    C_{\mathrm{rep}}=\max_{\ell\in\Lambda}C(\ell).
\end{equation*}
The fixed set $\Lambda$ contains nonzero spatial shifts with at least
half of the line overlapping. Constant traces and zero denominators
have undefined correlation. Values near one indicate similar responses
at separated locations. The score is read alongside the response peaks;
it alone does not establish a common spatial period.
Scores are averaged over the three coordinate directions within each
shape. Figure~\ref{fig:periodic-responses} and
Table~\ref{tab:figure2-correlation} keep surface-mass ranks 1, 10, and
the median separate; Figure~\ref{fig:component-profiles} also averages
these ranks for its model comparison.

\paragraph{(e) Query-normal sensitivity.}
Keep positions, labels, support features, and the posterior sampling
rule fixed. Change only query normals: retain them, set them to zero,
perturb and renormalize them, or randomly flip their signs.
Table~\ref{tab:query_normal} uses Gaussian perturbations with standard
deviation 0.2 and flip probability 0.5. Local selection, local mass,
and reconstruction CD measure dependence on this query cue.
Normal-free queries ignore these changes while support normals remain
available; the table reports this controlled comparison for A0, A2, and A3.

\paragraph{(f) Field orientation.}
Six queries at positive and negative offsets along the three coordinate
axes estimate the gradient at surface point $\mathbf x_k$:
\begin{equation*}
    [\mathbf g_k]_c=
       \frac{\hat f(\mathbf x_k+h\mathbf e_c)
                   -\hat f(\mathbf x_k-h\mathbf e_c)}{2h},
    \qquad c=1,2,3,
\end{equation*}
where $\mathbf e_c$ is a coordinate unit vector. The evaluation step
$h$ is independent of the training offset $\varepsilon$; the six probes
evaluate the complete gradient direction, while the three training
probes constrain variation along the reference normal.
For a fixed threshold $\eta>0$, the valid index set is
$\mathcal I_{\mathrm{valid}}=\{k:\lVert\mathbf g_k\rVert_2>\eta\}$.
Valid gradients yield unit vectors
$\widehat{\mathbf n}_k=\mathbf g_k/\lVert\mathbf g_k\rVert_2$.
Their angular error against unit reference normals $\mathbf n_k$ is
\begin{equation*}
    E_{\mathrm{angle}}=
    \frac{180}{\pi|\mathcal I_{\mathrm{valid}}|}
    \sum_{k\in\mathcal I_{\mathrm{valid}}}
       \arccos\!\left(\operatorname{clip}
       (\widehat{\mathbf n}_k^\top\mathbf n_k,-1,1)\right).
\end{equation*}
Lower error means closer agreement in surface orientation. For
$N_s$ evaluation points, the near-zero-gradient rate is
$1-|\mathcal I_{\mathrm{valid}}|/N_s$; an empty valid set has undefined
angular error. Paired angular-error differences use common valid
points and retain each model's invalid rate. Compared models share
surface samples and subsets. The in-the-wild profiles use 8192 points,
$h=0.002$, and $\eta=10^{-6}$.

\begin{table}[htbp]
\centering
\caption{Lookup for diagnostic names and symbols used in the figures
and tables. Coordinate units refer to the normalized geometry.
Fractions are displayed as percentages where indicated.}
\label{tab:profiling-metrics}
\begin{tabular}{@{}p{.26\textwidth}p{.16\textwidth}p{.33\textwidth}p{.15\textwidth}@{}}
\toprule
Metric & Symbol & Meaning & Unit \\
\midrule
Local selection & $L_{\mathrm{enc}}$ & \raggedright Queries selecting one of 64 neighbors & Fraction \\
Local mass & $P_{\mathrm{enc},k}$ & Attention on 64 nearest inputs & Fraction \\
Most-selected input & --- & \raggedright Largest share of queries selecting one input & Fraction \\
Token mass & $p_i$ & \raggedright Conditional share of a shape token & Fraction \\
Top-1\% share & $T_{1\%}$ & \raggedright Mass on the top 1\% of shape tokens & Fraction \\
Mass entropy & $H_{\mathrm{shape}}$ & Evenness of shape-token use & Unitless \\
Bank share & $G_{\mathrm{bank}}$ & \raggedright Mass on the decoder bank & Fraction \\
Read distance & $R_{\mathrm{shape}}$, $R_{\mathrm{top}}$ & \raggedright Distance from query to token's FPS center & Coordinate \\
Response spread & $S_i$, $S_{\mathrm{sel}}$, $S_{\mathrm{top}}$ & \raggedright Extent around response centroid & Coordinate \\
Shift correlation & $C_{\mathrm{rep}}$ & Largest nonzero-shift correlation & Unitless \\
Normal error & $E_{\mathrm{angle}}$ & \raggedright Gradient angle to surface normal & Degrees \\
\bottomrule
\end{tabular}
\end{table}

\clearpage
\subsection{Additional Baseline Attention Measurements}
\label{sec:measured-figure1}

\paragraph{Shared selection can coexist with local attention.}
The head-averaged encoder selects a nearby input for only 0.073\% of
queries, yet two individual heads have local-selection rates above
99\%. Local attention mass is 7.22\%, while the most-selected input
receives the dominant selection for 94.06\% of queries. These measures
capture different aspects of the same attention pattern: a common
strongest selection can coexist with useful local evidence in other heads.

\begin{table}[t]
\centering
\caption{Baseline attention diagnostics: means [95\% shape-bootstrap intervals]. Distances and spreads use canonical coordinates; lower values indicate greater spatial locality.}
\label{tab:figure1-measured}
\begin{tabular}{lcc}
\toprule
Statistic & Volume queries & Input-surface queries \\
\midrule
$100T_{1\%}$ & 19.50 [18.48, 20.61] & 9.83 [9.32, 10.39] \\
$H_{\mathrm{shape}}$ & 0.875 [0.871, 0.878] & 0.960 [0.959, 0.962] \\
$R_{\mathrm{shape}}\!\downarrow$ & 0.953 [0.937, 0.969] & 0.304 [0.288, 0.319] \\
$R_{\mathrm{top}}\!\downarrow$ & 1.030 [1.001, 1.061] & 0.393 [0.370, 0.418] \\
Uniform reference for $R_{\mathrm{top}}$ & 1.244 [1.224, 1.266] & 0.764 [0.726, 0.804] \\
$S_{\mathrm{top}}\!\downarrow$ (mass weighted) & 0.893 [0.882, 0.905] & 0.354 [0.338, 0.371] \\
Uniform-query spread & 1.029 [1.029, 1.029] & 0.619 [0.593, 0.645] \\
\bottomrule
\end{tabular}
\end{table}

\paragraph{Attention mass and spatial extent answer different questions.}
Table~\ref{tab:figure1-measured} compares decoder reads near the surface
and throughout the volume. The top 1\% of tokens receive 9.83\% of
surface attention and 19.50\% of volume attention. Surface responses
are more compact, but their mass remains concentrated. In the spatial
maps, each token's response is divided by its own peak; colors therefore
show where a token is read, rather than its absolute share of attention.
The FPS markers provide the geometric reference for reading those maps.

\subsection{Spatial Responses and Field Orientation}
\label{sec:measured-figure2}

\paragraph{Peak locations and response curves.}
Figure~\ref{fig:periodic-responses} follows tokens ranked first, tenth,
and median by surface attention. A maximum-response position shows where
a token is read most strongly; slices and line traces reveal whether it
is also retrieved at other locations. Table~\ref{tab:figure2-correlation}
reports spatial-shift correlations for lines through either the token's
FPS origin or its response peak. Positive correlations in both views
support spatially extended token responses.

\begin{table}[t]
\centering
\caption{Spatial-shift response correlation in the baseline, with 95\% confidence intervals. Tokens are ranked by surface attention mass; lines pass through either the FPS origin or the strongest response. Lower correlation indicates less spatial repetition; bold and underlined estimates mark the lowest and second-lowest per column.}
\label{tab:figure2-correlation}
\begin{tabular}{lcc}
\toprule
Surface mass rank & Lines through FPS origin\,$\downarrow$ & Lines through volume peak\,$\downarrow$ \\
\midrule
1 & \underline{0.493} [0.456, 0.531] & \underline{0.475} [0.434, 0.515] \\
10 & 0.548 [0.524, 0.575] & 0.518 [0.491, 0.547] \\
4097 (median) & \textbf{0.292} [0.268, 0.315] & \textbf{0.274} [0.250, 0.296] \\
\bottomrule
\end{tabular}
\end{table}

\paragraph{Field orientation.}
The baseline's mean angular error is $13.52^\circ$, while its mean
per-shape median and 90th percentile are $5.64^\circ$ and $37.12^\circ$.
The broad error distribution shows why accurate field values alone do
not fully characterize the reconstructed geometry. The angular profile
complements surface distance by measuring the direction of local field
variation.

\section{Additional Experiments}
\label{sec:extended-results}

The following results expand the main-text comparisons. A0 is standard
VecSet. A1 adds the
learnable decoder bank (LB), A2 adds distance bias (DB), A3 removes
encoder query normals (PQ), and A4 adds normal supervision (NS).
Each variant retains the preceding components.

Arrows indicate the preferred direction ($\uparrow$ higher, $\downarrow$
lower). \textbf{Bold} and \underline{underlined} scores mark the best
and second-best per column, with ties at the displayed precision sharing
a rank. Captions specify any separate comparison groups.

\subsection{Reconstructing Sharp Structure}

Global surface accuracy can hide errors around edges and narrow features.
Table~\ref{tab:tripo-sharp} therefore also reports sharp-point distance
(S-CD) and sharp-point coverage (S-F). Lower S-CD means that the
reconstruction lies closer to ground-truth sharp features; higher S-F
means that more of those features are recovered within the threshold.
The cumulative variants improve both measures, complementing the
reconstruction gains in the main text.

\begin{table}[htbp]
\centering
\caption{Sharp-structure reconstruction for A0--A4 at 4096 tokens.
CD and sharp-point distance S-CD are scaled by $10^3$;
F1@.001 and sharp-point coverage S-F@.01 are percentages.}
\label{tab:tripo-sharp}
\footnotesize
\begin{tabular}{lrrrrr}
\toprule
Model & Tokens & CD\,$\downarrow$ & F1@.001\,$\uparrow$
& S-CD\,$\downarrow$ & S-F@.01\,$\uparrow$ \\
\midrule
A0: VecSet & 4096 & 3.723 & 35.15 & 3.049 & 97.30 \\
A1: + LB & 4096 & 2.960 & 42.70 & 2.803 & 98.30 \\
A2: + DB & 4096 & 2.809 & 45.99 & 2.759 & 98.44 \\
A3: + PQ & 4096 & \underline{2.723} & \underline{46.71} & \underline{2.480} & \underline{99.11} \\
A4: + NS & 4096 & \textbf{2.526} & \textbf{49.12} & \textbf{2.281} & \textbf{99.50} \\

\bottomrule
\end{tabular}
\end{table}

\subsection{Storage and Inference Cost}

The shared decoder bank changes computation without enlarging the
stored shape representation. For $M$ tokens of width $C_z$, storage
remains proportional to $MC_z$, while the decoder processes $M+K$
tokens. Table~\ref{tab:efficiency} shows that the reconstruction gains
come with a modest increase in latent-refinement time relative to VecSet.
Encoding, refinement, and spatial queries are shown separately to make
this tradeoff visible.

\begin{table}[htbp]
\centering
\caption{Representation storage and inference cost. KiB is stored latent
size and Mem. is peak memory in GiB. Encode constructs the latent code;
Decode refines it; KV prepares decoder keys and values; Query evaluates
4096 spatial locations. Times are in milliseconds and exclude mesh
extraction. Blocks separate cumulative variants and full-scale models
at each token budget; memory and times are ranked within each block.}
\label{tab:efficiency}
\footnotesize
\setlength{\tabcolsep}{4pt}
\begin{tabular}{lrrrrrrr}
\toprule
Model & Tokens & KiB & Mem.\,$\downarrow$
& Encode\,$\downarrow$ & Decode\,$\downarrow$ & KV\,$\downarrow$
& Query\,$\downarrow$ \\
\midrule
A0: VecSet & 4096 & 512 & \textbf{3.754} & \textbf{1273.9} & \textbf{53.96} & \textbf{0.466} & 2.189 \\
A1: + LB & 4096 & 512 & \underline{3.755} & 1276.5 & \underline{56.33} & \underline{0.484} & 2.254 \\
A2: + DB & 4096 & 512 & \underline{3.755} & \underline{1275.3} & 56.34 & 0.495 & \textbf{2.146} \\
A3: + PQ & 4096 & 512 & \underline{3.755} & 1275.6 & 56.46 & 0.494 & \underline{2.148} \\
A4: + NS & 4096 & 512 & \underline{3.755} & 1275.5 & 56.49 & 0.497 & 2.175 \\
\midrule
VecSet VAE$^\dagger$ & 4096 & 512 & \textbf{2.319} & \underline{649.7} & \underline{53.99} & \textbf{0.465} & 3.806 \\
LATTICE$^\dagger$ & 4096 & 512 & \underline{2.320} & \textbf{312.5} & \textbf{53.87} & \textbf{0.465} & \underline{2.187} \\
\method & 4096 & 512 & \underline{2.320} & 649.8 & 56.32 & \underline{0.493} & \textbf{2.149} \\
\midrule
VecSet VAE$^\dagger$ & 8192 & 1024 & \textbf{3.802} & \underline{2544.2} & \underline{146.04} & \underline{0.898} & 4.958 \\
LATTICE$^\dagger$ & 8192 & 1024 & \underline{3.803} & \textbf{670.3} & \textbf{145.82} & \textbf{0.897} & \underline{3.066} \\
\method & 8192 & 1024 & \underline{3.803} & 2546.2 & 150.84 & 0.932 & \textbf{2.927} \\

\bottomrule
\end{tabular}
\end{table}

\subsection{Changes Across Cumulative Variants}

Table~\ref{tab:tripo-paired} measures the change between successive
variants on the same objects. The decoder bank gives the clearest
initial CD reduction. Distance bias and normal-free queries further
improve the point estimates, while their reconstruction intervals
include zero. Their spatial effects are clearer in the attention
profiles below. A4 improves reconstruction while increasing angular
error, reinforcing the distinction between surface fit and field orientation.

\begin{table}[htbp]
\centering
\caption{Successive variant minus preceding variant, with 95\%
confidence intervals across shapes. CD is scaled by $10^3$, F1 changes
are percentage points, and normal-error changes are degrees. Negative
CD changes and positive F1 changes indicate better reconstruction.
Emphasis ranks point estimates of the changes, not statistical significance.}
\label{tab:tripo-paired}
\footnotesize
\setlength{\tabcolsep}{4pt}
\begin{tabular}{lccc}
\toprule
Comparison & $\Delta$ CD\,$\downarrow$ & $\Delta$ F1@.001\,$\uparrow$
& $\Delta E_{\mathrm{angle}}$\,$\downarrow$ \\
\midrule
A0 $\to$ A1 & \textbf{-0.763} [-0.958, -0.455] & \textbf{7.55} [4.83, 9.47] & \textbf{-0.91} [-1.02, -0.81] \\
A1 $\to$ A2 & -0.150 [-0.408, 0.055] & \underline{3.29} [0.02, 5.76] & -0.31 [-0.35, -0.27] \\
A2 $\to$ A3 & -0.087 [-0.331, 0.107] & 0.72 [-2.11, 3.53] & \underline{-0.80} [-0.89, -0.71] \\
A3 $\to$ A4 & \underline{-0.196} [-0.409, -0.007] & 2.42 [0.24, 5.63] & 3.66 [3.26, 4.07] \\

\bottomrule
\end{tabular}
\end{table}

\subsection{Spatial Retrieval and Query-Normal Sensitivity}

\paragraph{The distance bias gives retrieval a spatial reference.}
Table~\ref{tab:attention_profile} separates attention near the surface
from attention throughout the volume. With distance bias, shape-token
reads become shorter and their spatial responses more compact. Turning
the bias off in A2 increases surface read distance from 0.035 to 0.607
and angular error from $11.61^\circ$ to $29.81^\circ$. The anchors and
latent code are unchanged, showing that the learned decoder actively
uses the spatial prior during retrieval.

\begin{table}[htbp]
\centering
\caption{Decoder attention profiles. $T_{1\%}$ is the attention share
of the most-read 1\% of shape tokens; $H_{\mathrm{shape}}$ measures how
evenly shape tokens are read. $G_{\mathrm{bank}}$
is the shared bank's attention share. $R_{\mathrm{shape}}$ and
$R_{\mathrm{top}}$ are read distances for all shape tokens and the
most-read 1\%, respectively; $S_{\mathrm{sel}}$ measures spatial response
spread. Shares are percentages; distances use normalized coordinates.
Distance and spread are ranked separately for surface and volume probes.
Mass shares and entropy are descriptive, with no preferred direction.}
\label{tab:attention_profile}
\footnotesize
\begin{tabular}{llrrrrrr}
\toprule
Model & Probe & $T_{1\%}$ & $H_{\mathrm{shape}}$ & $G_{\mathrm{bank}}$
& $R_{\mathrm{shape}}\!\downarrow$ & $R_{\mathrm{top}}\!\downarrow$
& $S_{\mathrm{sel}}\!\downarrow$ \\
\midrule
A0 & Surface & 1.95 & 0.995 & 0.00 & 0.205 & 0.217 & 0.343 \\
A0 & Volume & 5.00 & 0.985 & 0.00 & 0.874 & 1.197 & 0.921 \\
A1 & Surface & 2.66 & 0.992 & 31.19 & 0.175 & 0.320 & 0.450 \\
A1 & Volume & 3.18 & 0.986 & 38.43 & 0.834 & 0.900 & 0.926 \\
A2 & Surface & 2.46 & 0.990 & 54.27 & \textbf{0.035} & \textbf{0.028} & \textbf{0.041} \\
A2 & Volume & 2.00 & 0.996 & 97.04 & \textbf{0.079} & \textbf{0.085} & \textbf{0.098} \\
A2 (bias off) & Surface & 1.89 & 0.996 & 29.50 & 0.607 & 0.659 & 0.569 \\
A2 (bias off) & Volume & 1.91 & 0.996 & 30.87 & 1.143 & 1.201 & 1.028 \\
A3 & Surface & 2.38 & 0.991 & 53.17 & 0.039 & \underline{0.029} & 0.044 \\
A3 & Volume & 2.32 & 0.994 & 96.40 & 0.093 & 0.105 & 0.108 \\
A4 & Surface & 2.48 & 0.990 & 48.69 & \underline{0.036} & \textbf{0.028} & \underline{0.042} \\
A4 & Volume & 2.23 & 0.994 & 96.24 & \underline{0.081} & \underline{0.092} & \underline{0.103} \\

\bottomrule
\end{tabular}
\end{table}

\paragraph{Normal-free queries remove dependence on an unavailable cue.}
Zeroing query normals raises A0's CD from 4.037 to 16.414 and A2's from
3.026 to 3.747 (Table~\ref{tab:query_normal}). In contrast, A3 retains
CD 2.776 and 100\% local selection under every perturbation. The encoder
still observes support normals, but its queries no longer require them.
This preserves orientation information while avoiding a query feature
that arbitrary decoder locations cannot provide.

\begin{table}[htbp]
\centering
\caption{Effect of changing only encoder query normals. A0 is VecSet;
A2 includes the bank and distance bias; A3 additionally removes query
normals. Local selection $L_{\mathrm{enc}}$ and local attention mass
$P_{\mathrm{enc},k}$ are percentages; CD is scaled by $10^3$.
Positions and support features remain fixed. Scores are ranked across
models under the same query-normal condition.}
\label{tab:query_normal}
\footnotesize
\begin{tabular}{llrrr}
\toprule
Model & Query normals & Local selection\,$\uparrow$ & Local mass\,$\uparrow$
& CD\,$\downarrow$ \\
\midrule
A0 & Clean & \underline{0.076} & 3.05 & 4.037 \\
A0 & Zeroed & \underline{0.052} & 3.04 & 16.414 \\
A0 & Perturbed & \underline{0.078} & 3.05 & 4.382 \\
A0 & Random flips & \underline{0.076} & 2.98 & 5.759 \\
A2 & Clean & 0.043 & \underline{9.77} & \underline{3.026} \\
A2 & Zeroed & 0.043 & \underline{9.87} & \underline{3.747} \\
A2 & Perturbed & 0.046 & \underline{9.76} & \underline{3.248} \\
A2 & Random flips & 0.042 & \underline{9.47} & \underline{4.022} \\
A3 & Clean & \textbf{100.000} & \textbf{53.82} & \textbf{2.776} \\
A3 & Zeroed & \textbf{100.000} & \textbf{53.82} & \textbf{2.776} \\
A3 & Perturbed & \textbf{100.000} & \textbf{53.82} & \textbf{2.776} \\
A3 & Random flips & \textbf{100.000} & \textbf{53.82} & \textbf{2.776} \\

\bottomrule
\end{tabular}
\end{table}

\newpage
\subsection{Surface Accuracy and Field Orientation}

Table~\ref{tab:normal_error} compares the field gradient with the
reference surface normal, both over the whole surface and near regions
of high curvature. A3 improves orientation relative to the baseline,
whereas A4 combines better reconstruction with a larger angular error.
The same distinction appears near high-curvature features: the error
rises from $16.84^\circ$ to $18.14^\circ$ between A3 and A4.
A directional slope constraint therefore complements surface fitting,
but does not by itself constrain every component of the field gradient.

\begin{table}[htbp]
\centering
\caption{Reconstruction and field orientation. Angular errors are in
degrees; the high-curvature subset contains points with the largest local
normal variation. CD is scaled by $10^3$ and sharp-point coverage S-F
is a percentage. Lower angular error indicates closer agreement between
predicted gradients and reference normals.}
\label{tab:normal_error}
\footnotesize
\begin{tabular}{lrrrr}
\toprule
Model & Angular error\,$\downarrow$ & High-curvature error\,$\downarrow$
& CD\,$\downarrow$ & S-F@.01\,$\uparrow$ \\
\midrule
A0: VecSet & 12.83 & 21.01 & 3.723 & 97.30 \\
A1: + LB & 11.92 & 19.09 & 2.960 & 98.30 \\
A2: + DB & \underline{11.61} & 18.50 & 2.809 & 98.44 \\
A3: + PQ & \textbf{10.82} & \textbf{16.84} & \underline{2.723} & \underline{99.11} \\
A4: + NS & 14.48 & \underline{18.14} & \textbf{2.526} & \textbf{99.50} \\

\bottomrule
\end{tabular}
\end{table}

\clearpage
\section{Additional Qualitative Results}
\label{sec:qualitative-gallery}

\begin{figure}[htbp]
\centering
\includegraphics[page=1,width=\textwidth]{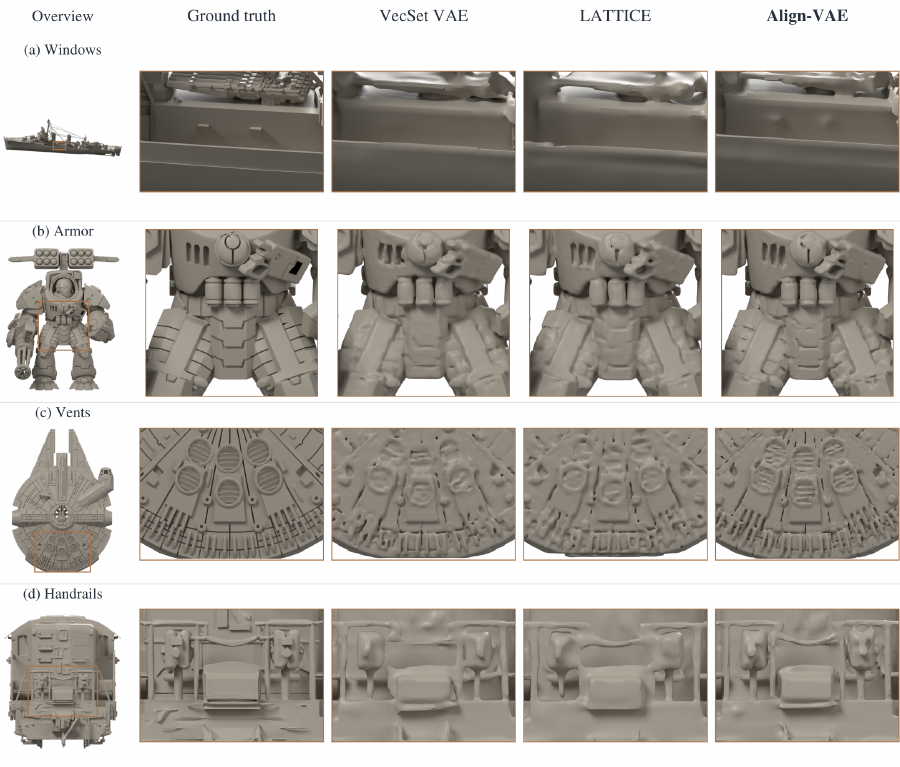}
\caption{Additional qualitative results on selected shapes at 8192 tokens:
(a) hull windows, (b) waist canisters and armor plates, (c) vents and rim struts,
and (d) handrails and fixtures. Ground-truth overviews locate each detail.
All methods share the camera, material, lighting, and crop within each row,
with details rendered directly at higher resolution.}
\label{fig:reconstruction-gallery}
\end{figure}

\begin{figure}[!ht]
\centering
\includegraphics[page=2,width=\textwidth]{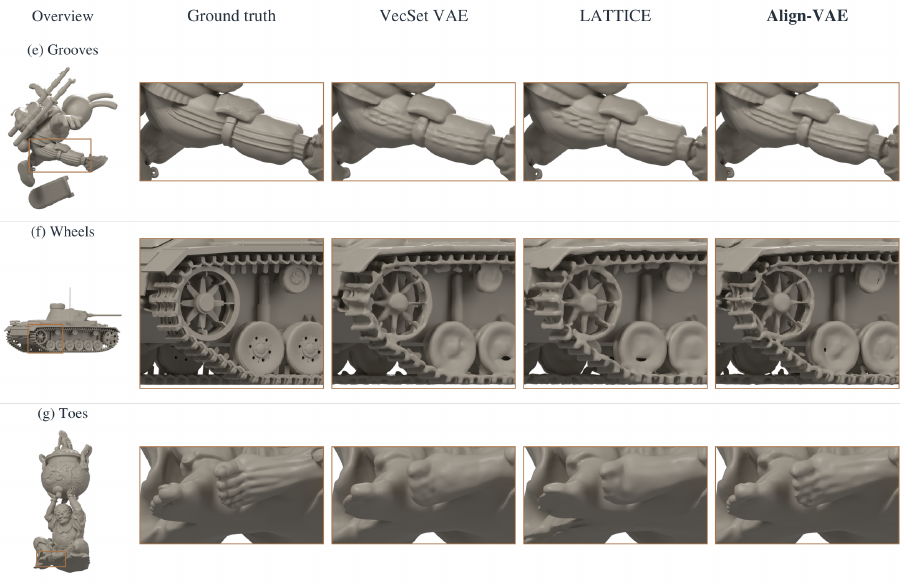}
\caption{Additional qualitative results on selected shapes at 8192 tokens
(continued): (e) trouser grooves, (f) wheel spokes and rim, and (g) individual
toes. The same rendering protocol is used as in
Figure~\ref{fig:reconstruction-gallery}, preserving the size of the enlarged
details and the matched views across methods.}
\label{fig:reconstruction-gallery-continued}
\end{figure}

Figures~\ref{fig:reconstruction-gallery} and~\ref{fig:reconstruction-gallery-continued}
compare seven additional selected shapes at 8192 tokens. The enlarged views
illustrate local geometric detail without implying lower whole-shape
reconstruction error.

\clearpage

\clearpage

\end{document}